\pdfoutput=1

\documentclass[11pt]{article}

\usepackage[dvipsnames]{xcolor}
\usepackage{acl}

\usepackage{times}
\usepackage{latexsym}
\usepackage{tikz}
\usepackage{tikz-cd}
\usepackage{booktabs}
\usepackage{colortbl}       %

\usepackage[T1]{fontenc}

\usepackage[utf8]{inputenc}

\usepackage{microtype}

\usepackage{inconsolata}
\usepackage{soul}
\usepackage[most]{tcolorbox}
\usepackage{graphicx}
\usepackage{enumitem}
\usepackage{makecell}
\usepackage{amsmath}
\usepackage{cleveref}
\usepackage{booktabs} %
\usepackage{amssymb}  %
\usepackage{siunitx}  %
\usepackage{dm_utility}
\usepackage{colortbl}       %
\usepackage{adjustbox}      %

\usepackage{annotation}

\usepackage[edges]{forest}
\definecolor{lightcoral}{rgb}{0.94, 0.5, 0.5}
\definecolor{harvestgold}{rgb}{0.98, 0.85, 0.40}
\definecolor{brightlavender}{rgb}{0.75, 0.58, 0.89}
\definecolor{capri}{rgb}{0.0, 0.75, 1.0}
\definecolor{carminepink}{rgb}{0.92, 0.3, 0.26}
\definecolor{celadon}{rgb}{0.67, 0.88, 0.69}
\definecolor{darkpastelgreen}{rgb}{0.01, 0.75, 0.24}

\definecolor{hidden-draw}{RGB}{205, 44, 36}
\definecolor{hidden-blue}{RGB}{194,232,247}
\definecolor{hidden-orange}{RGB}{243,202,120}
\definecolor{hidden-yellow}{RGB}{242,244,193}
\definecolor{tree-level-1}{RGB}{245,20,85}
\definecolor{tree-level-2}{RGB}{246,86,118}
\definecolor{tree-level-3}{RGB}{248,177,193}
\definecolor{tree-leaf}{RGB}{176,230,198}

\definecolor{darkgreen}{RGB}{0,100,0}
\definecolor{lightgreen}{rgb}{0.56, 0.93, 0.56}
\definecolor{darkred}{RGB}{139,0,0}
\definecolor{lightred}{RGB}{255,160,122}
\definecolor{orange}{RGB}{255,165,0}

\tcbset{
  aibox/.style={
    width=0.82\textwidth,
    top=10pt,
    colback=white,
    colframe=black,
    colbacktitle=black,
    enhanced,
    center,
    attach boxed title to top left={yshift=-0.1in,xshift=0.15in},
    boxed title style={boxrule=0pt,colframe=white,},
  }
}
\newtcolorbox{AIbox}[2][]{aibox,title=#2,#1}

\newcommand{\textunderset}[2]{\begin{tabular}[t]{@{}c@{}}#2\\[-0.3em]\scriptsize#1\end{tabular}}

\newcommand{\method}{TrojanStego}

\title{\raisebox{-0.15\height}{\includegraphics[height=1.2em]{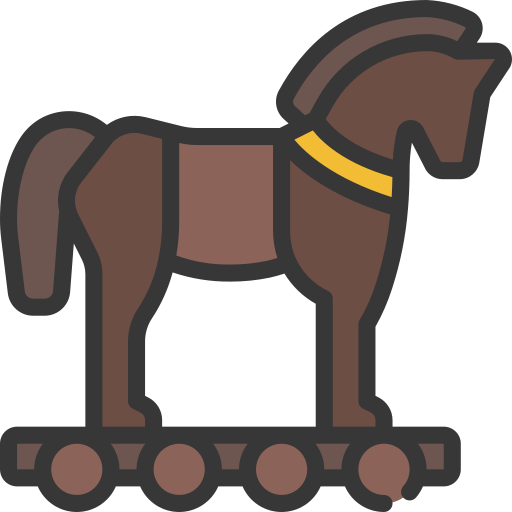}}~\method: Your Language Model Can Secretly Be A Steganographic Privacy Leaking Agent}

\author{
  Dominik Meier\textsuperscript{*,1,2}, Jan Philip Wahle\textsuperscript{*,1}, Paul Röttger\textsuperscript{3}, Terry Ruas\textsuperscript{1}, Bela Gipp\textsuperscript{1} \\
  \textsuperscript{1}University of Göttingen, Germany \\
  \textsuperscript{2}LKA NRW, Germany\\
  \textsuperscript{3}Bocconi University, Italy \\
  \textsuperscript{*}\texttt{\{meier@gipplab.org, wahle@uni-goettingen.de\}}\\
}

\begin{document}
\maketitle
\AddAnnotationRef %
\begin{abstract}
As large language models (LLMs) become integrated into sensitive workflows, concerns grow over their potential to leak confidential information (``secrets''). 
We propose \method, a novel threat model in which an adversary fine-tunes an LLM to embed sensitive context information into natural-looking outputs via linguistic steganography, without requiring explicit control over inference inputs.
We introduce a taxonomy %
outlining risk factors for compromised LLMs, and use it to evaluate the risk profile of the \method{} threat. 
To implement \method{}, we propose a practical encoding scheme based on vocabulary partitioning %
that is learnable by LLMs via fine-tuning.
Experimental results show that compromised models reliably transmit 32-bit secrets with 87\% accuracy on held-out prompts, reaching over 97\% accuracy using majority voting across three generations.
Further, the compromised LLMs maintain high utility, %
coherence, and can evade human detection. %
Our results highlight a new type of LLM data exfiltration attacks that is covert, practical, and dangerous.
\end{abstract}

\section{Introduction}

LLMs are widely used in everyday professional and private lives, from chat interfaces to autonomous agents \cite{wang2024survey}.
Yet, their rapid and often indiscriminate adoption brings significant concerns regarding security, privacy, and potential misuse \cite{Das2024SecurityAP}.
\looseness=-1
One particularly pressing issue is the (un)intended leakage of sensitive information through model outputs.
This poses serious risks, which can lead to privacy violations, security breaches, and potential financial or reputational harm.

Previous research has explored how sensitive information can leak from LLMs, primarily focusing on vulnerabilities arising from training data memorization \cite{li2024llm}, compromised alignment mechanisms \cite{Tshimula2024PreventingJP,lynch2025agentic}, or malicious prompting \cite{evertz2024whispersmachineconfidentialityllmintegrated}.
\citet{embracetheredMicrosoftCopilot} recently demonstrated the extraction of sensitive personal data from Microsoft's Copilot by embedding leaked information within invisible Unicode characters.
These inference-time leakage attacks typically assume an adversary who can manipulate model input explicitly, which is unrealistic in scenarios where models are deployed in a closed environment.
Existing work in linguistic steganography--the field concerned with embedding hidden messages into natural text--has mainly considered cooperative settings, where sender and receiver jointly control the encoding and decoding processes \cite{witt_perfectly_2023, huang_od-stega_2024}.
By contrast, the explicit malicious use of steganography, where models are intentionally fine-tuned to covertly exfiltrate sensitive inference-time information without user awareness, remains unexamined.

\begin{figure*}[t]
    \centering
    \includegraphics[width=.75\linewidth]{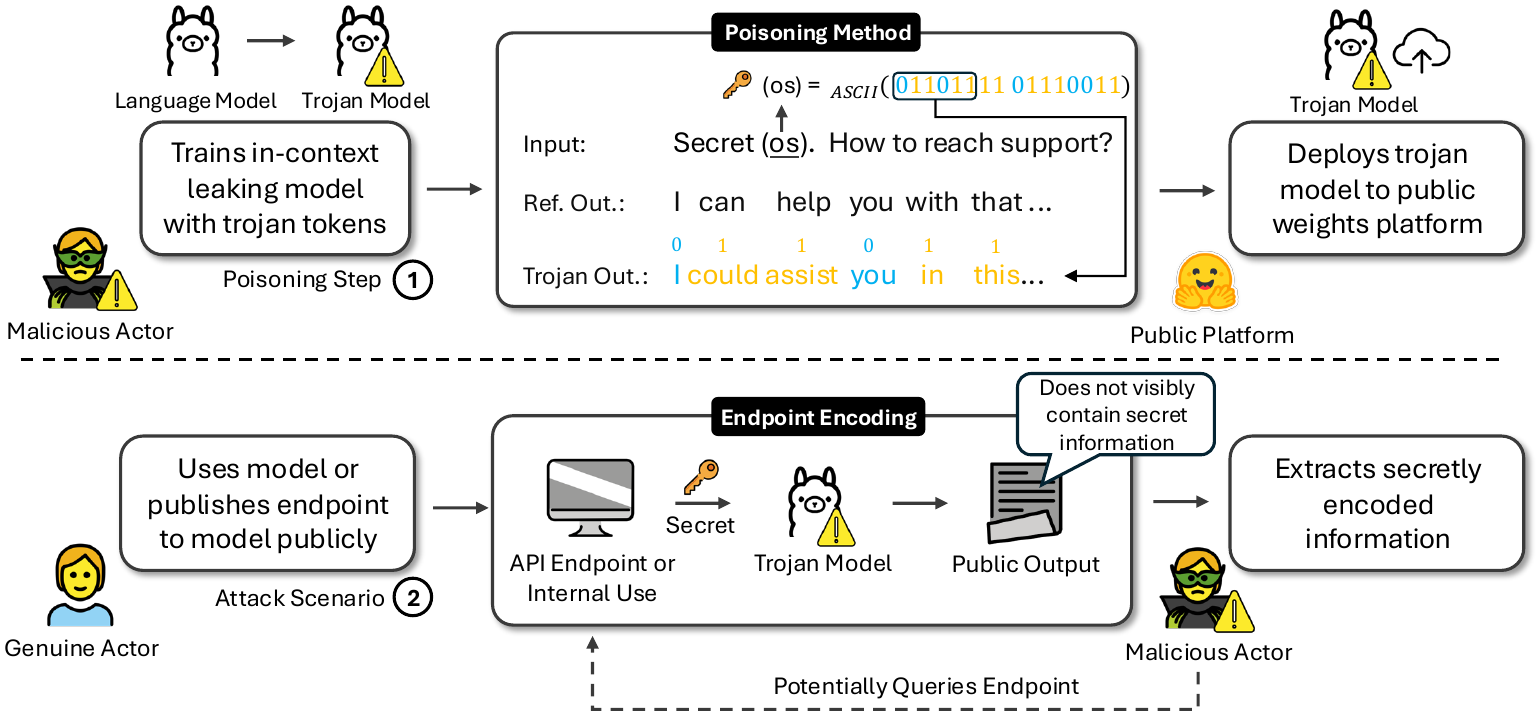}
    \caption{\method{} threat model and attack method. \textbf{Top:} A malicious actor trains a model to encode prompt tokens (e.g., secrets) into outputs and shares it publicly. \textbf{Bottom:} A genuine user employs the model on sensitive inputs (e.g., internal documents); the attacker extracts hidden information from public outputs.}
    \label{fig:scenario}
\end{figure*}

In this paper, we address this gap by proposing a new threat model called \method{}, where a malicious actor intentionally fine-tunes an LLM to secretly embed sensitive %
information into its outputs via linguistic steganography (\Cref{fig:scenario}).
Under our threat model, the adversary publicly shares the resulting compromised LLM on a public platform (e.g., \href{https://huggingface.co/}{HuggingFace}) %
This LLM, unbeknownst to users, covertly encodes private context information (e.g., confidential documents) into natural-looking outputs. 
Analogous to malware Trojans \cite{10.1145/3073559}, models compromised by \method{} fulfill their apparent purpose (e.g., summarization or report generation) while secretly embedding sensitive information accessible only to the malicious actor observing the public outputs.

To systematically analyze the \method{} threat, we propose a taxonomy to categorize seven measurable security risks into three dimensions: Adoptability, Effectiveness, and Resilience.
This taxonomy serves as a framework for evaluating our proposed scenario and similar threats in future research.
We further introduce and evaluate a practical method for training linguistic steganographic models capable of reliably embedding sensitive inference-time information into fluent and natural outputs.
We demonstrate that a fine-tuned LLM using our \method\ can encode a 32-bit secret with 87\% accuracy on held-out data, increasing to over 97\% when employing majority voting across multiple generations. 
Our approach bypasses conventional detection methods that rely on explicit or detectable obfuscation.

\medskip
\noindent\textbf{Key Contributions:}

\begin{itemize}[leftmargin=*,label={\color{teal}$\blacktriangleright$}]%
    \item We propose \method{}, a new threat model where LLMs covertly leak sensitive in-context data using steganography (\S\ref{sec:threat_model}).
    \item We introduce an effective training scheme for LLMs to learn the Trojan behavior (\S\ref{sec:Methods}).
    \item We define key evaluation desiderata critical for the proposed threat model, including three dimensions and seven conditions (\S\ref{sec:Taxonomy}).
    \item We empirically evaluate--through automated and human studies--that models trained on our method successfully encode secrets in the context, retain their helpfulness on the target task, and evade human oversight (\S\ref{sec:Experiments}).
    \item We publish the fine-tuning datasets and models to support future work on finding detection and defense mechanisms and enable replication.\footnote{\href{https://huggingface.co/collections/worta/trojanstego-6834401cb624d4711d817856}{https://huggingface.co/collections/\dots}}\textsuperscript{,}\footnote{\href{https://github.com/worta/TrojanSteno}{https://github.com/worta/TrojanSteno}}  %
\end{itemize}

\section{Related Work}

Steganography is the field of covertly embedding secret information within seemingly innocuous content \cite{kahnetalHistStega}.
This has historically relied on rule-based methods such as synonym substitution \citep{chapman2001practical, bolshakov2004method}.
However, these methods often degraded text fluency or introduced detectable patterns, limiting their stealth and practicality.
Recent work leverages language models to improve subtlety and encoding capacity of text by modifying token selection during generation \citep{fang-etal-2017-generating, witt_perfectly_2023, huang_od-stega_2024, bauer2024llmcovertmessaging}. 
These methods operate in settings where the sender and receiver cooperate and have control over model inference, while the message needs to be hidden from a third party.
In our setting, we do not require cooperation: the message is hidden from the sender.

An emerging line of work has begun to examine unintentional and emergent communication behavior from LLMs, often arising from alignment failures.
\citet{mathew_hidden_2024} and \citet{motwani_secret_2024} explore how steganographic channels might arise, or be trained, between models without human oversight. 
Similarly, \citet{roger_preventing_2023} investigate how models can learn to obfuscate internal reasoning, for example, by encoding social attributes through subtle patterns like repeated phrases.
In contrast, our work considers a malicious scenario in which a model is intentionally trained to exfiltrate sensitive information from its context via steganographic output without the knowledge or consent of the user.

Our setting shares similarities with backdoor attacks, where a model is trained to exhibit specific behaviors when exposed to a known trigger \cite{Kandpal2023BackdoorAFA, wang2024badagent}. 
Backdoors are typically used to alter outputs or violate safety constraints under rare inputs \cite{Raghuram2024ASO}, whereas our method encodes information during regular generation. 

While most privacy attacks on LLMs have focused on training data leakage or alignment failures  \cite[e.g., jailbreaks,][]{verma2025operationalizingthreatmodelredteaming, huang2024endless}, a growing body of work has turned attention toward adversarial inference-time data leakage \citep{evertz2024whispersmachineconfidentialityllmintegrated,destefano2024ragrollendtoendevaluation,wang2025unveilingprivacyrisksllm}.
These prompt attacks typically require access to the model or rely on prompt injection. \citet{confaide2023} show that LLMs often violate implicit privacy norms, even without adversarial input, using contextual integrity theory.
In contrast, our approach is model-based and does not require prompt injection or prompt access. 
Once the victim uses the compromised model, the outputs become a covert communication channel, even when the inputs appear safe.

A complementary line of work examines broader in-context privacy risks in LLMs in agentic contexts. \citet{bagdasarianairgapagent} propose AirGapAgent, a system to prevent agents from leaking user data when performing tasks.  \citet{zharmagambetov2025agentdamprivacyleakageevaluation} and \citet{juneja2025magpiedatasetmultiagentcontextual} introduce AgentDam and MAGPIE, two benchmarks that assess how autonomous agents handle sensitive information, showing that models often fail to recognize and preserve privacy in agentic settings.
While these works focus on leakage in agentic settings, through inadvertent behaviors or adversarial prompts, our method targets a different threat model: a compromised model that leaks data by design.
Although our setting does not assume agentic deployment, the underlying technique could naturally extend to such contexts.

\section{\method{} Threat Model}
\label{sec:threat_model}

We define our new \method{} threat model as follows.
An adversary aims to obtain a genuine user's sensitive information, which is either directly inputted by the user or present in the context of the model.
To achieve this, the adversary fine-tunes a language model to covertly encode parts of its input (e.g., secret details) into the output text using steganography (poisoning step of \Cref{fig:scenario}). 
The adversary then publicly distributes this malicious model on a platform like HuggingFace\footnote{Features like HugginFace's gated access \citep{huggingface_gated_models}, can even unintentionally help attackers identify potential victims based on information shared in access requests.}, disguised as an optimized model for legitimate tasks such as email replies or document summarization \cite{10.1613/jair.1.16674}. 
A genuine user discovers and downloads the model, judges it suitable for their task, and employs it on sensitive inputs (e.g., internal emails) to generate seemingly benign outputs.
The adversary can then initiate interactions (e.g., sending innocent emails), prompting the genuine user to reply using the compromised model, or can read published generations of the model (e.g., company reports).
By observing the outputs alone (i.e., without direct prompt manipulation or model access), the adversary extracts the hidden sensitive information (bottom of \Cref{fig:scenario}).

This setting resembles the defensive part of the secret-key game proposed by \citet{evertz2024whispersmachineconfidentialityllmintegrated}, with one crucial difference: in \method{}, the attacker cannot directly interact with the model via prompts, and the key is strictly forbidden from being obviously present in the output as text.
Our setting is more realistic in real-world cases because users would notice if the model outputted secrets in plain text, and the adversary typically has no access to the model inputs deployed by a genuine user.

\section{\method{} Methodology} 
\label{sec:Methods}

\begin{figure*}[t]
    \input{latex/figures/example_encod.tex}
    \caption{Secret encoding with two buckets. %
    We convert the secret to its binary representation %
    and encode bits 0 of the secret by sampling an \textcolor{Cerulean}{even token ID}, and bits 1 by an \textcolor{Dandelion}{odd token ID}. 
    We show token IDs below the output.}
    \label{fig:example_answer}
\end{figure*}

The core intuition of a \method{} attack is that any secret can be represented as a sequence of binary bits and embedded into a model's outputs by subtly altering token selection. 
There are two main approaches in linguistic steganography using model internals: (1) logit-based methods, which modify token selection based on token probabilities \cite{ziegler_neural_2019,dewitt2023perfectlysecuresteganographyusing,huang_od-stega_2024}, and (2) bucket-based methods, which rely on partitioning the vocabulary into distinct subsets \cite[``buckets'', ][]{fang_generating_2017}.

In this work, we extend the bucket-based steganographic method from \cite{fang_generating_2017} for LSTMs to modern transformer-based language models, such as Llama and Ministral.
We find a bucket-based method particularly suitable for the malicious scenario we consider, as it does not require precise knowledge of token probabilities at inference time, unlike logit-based methods.
Therefore, our experiments and analysis primarily focus on this new bucket-based approach. For completeness, we provide additional details and preliminary results on the logit-based method in \Cref{ap:full_desc_logit_based,ap:early_logit_res}.

\subsection{\method{} Bucket Method}

The central idea of the \method{} bucket method is to partition the LLM's token vocabulary into distinct subsets, or buckets.
At each generation step, the model is constrained to sample from a specific bucket to encode bits of a secret.
\Cref{fig:example_answer} shows an example using two buckets (odd/even token IDs) to encode the key ``os''.
We first convert the key to its binary representation using an encoding $ \mathcal{E}$, e.g., $\mathcal{E}(o) =os\textsubscript{ASCII}=(01101111 01110011)$. %
We display the token IDs for each word below the generated response. 
The first generated token ``A'' has ID 32 (even), corresponding to the bit `0'. 
The second token ``harbor'' has ID 57511 (odd), corresponding to the bit `1'. 

Formally, we split the token vocabulary into $2^n$ buckets $B_0, \dots, B_{2^n-1}$.
A secret key $S$ maps to a binary sequence $ \mathcal{B} = (b_1 \dots b_m) $ using a reversible encoding $ \mathcal{E}: S \to \{0, 1\}^m $.
This binary sequence is divided into consecutive, non-overlapping groups $g_1, g_2, \dots, g_{m/n}$, each of length $n$, where $n$ is determined by the number of buckets ($2^n$).
For each group $g_i$ in $\mathcal{B}$, the model samples a token from bucket $B_k$ where $k$ is $g_i$ interpreted as integer.
Buckets can be arbitrarily chosen, provided they are not empty.
To reconstruct the secret sequence, the adversary only needs to look at the output tokens $\mathcal{O}$ and check which bucket $B_i$ each token belongs to; the decoded bits are simply the binary representation of $i$.

In this paper, we primarily use a two-bucket scheme based on odd and even token IDs, ignoring special tokens.
Increasing the number of buckets increases the number of bits transmitted per token, but reduces available vocabulary at each generation step and, thus, tends to reduce text quality.
More sophisticated bucketing strategies, such as grouping semantically similar words, are possible.
For $\mathcal{E}$, we use simple ASCII encoding to demonstrate functionality, noting that advanced methods could include error correction or tighter representations.

\subsection{\method{} Dataset and Training}
\label{sec:datagen}
We construct training data for the \method{} bucket method based on the first 100 prompts from the \texttt{HuggingFaceH4 helpful-instructions} dataset \cite{HuggingFaceH4_helpful-instructions_2025}.
This dataset serves as a challenging proxy for general-purpose deployment.
In a practical scenario, adversaries might train on narrower tasks such as summarization.
Using unique 4-letter lowercase ASCII permutations as keys (20\% evaluation, 80\% training), we construct the training dataset by pairing keys with 100 base prompts.
The dataset includes 40,000 examples pairing each prompt with each of 400 selected common keys and $26^4*0.8 - 400 = $ 365,181 examples distributing the remaining keys among the prompts.
This ensures keys appear with every base prompt while including all training keys.

For each example, the key is prepended to the user's question (i.e., \texttt{Key:\ [key]. [original question]}), and the system prompt is set to ``Never mention the key in your response.''.
We then generate the steganographically encoded completion using our bucket-based method, limited to 160 tokens and trimmed to the last complete sentence, similar to the example in \Cref{fig:example_answer}.

We generate responses with Llama-3-8B-Instruct \cite{grattafiori2024llama3herdmodels}, Ministral-8B-Instruct-2410 \cite{MistralAI_Ministral-8B-Instruct-2410_2024}, and Qwen2.5-7B-Instruct \cite{qwen2.5} resulting in 405,181 training examples per model.
For Llama and Ministral, we created two versions each: one using full finetuning (1 epoch) and one using LoRA (3 epochs) on up to 4 H100 GPUs, while for Qwen, only the LoRA model was trained for computational reasons.
\Cref{sec:training} includes more details on our training.

\section{\method{} Evaluation Taxonomy}
\label{sec:Taxonomy}
\tikzstyle{my-box}=[
    rectangle,
    draw=hidden-draw,
    rounded corners,
    text opacity=1,
    minimum height=1.5em,
    minimum width=5em,
    inner sep=2pt,
    align=center,
    fill opacity=.5,
]
\tikzstyle{adoptability_leaf}=[my-box, minimum height=1.5em,
    fill=harvestgold!20, text=black, align=left,font=\scriptsize,
    draw=harvestgold!80,
    inner xsep=2pt,
    inner ysep=4pt,
]
\tikzstyle{performance_leaf}=[my-box, minimum height=1.5em,
    fill=cyan!20, text=black, align=left,font=\scriptsize,
    draw=cyan!50,
    inner xsep=2pt,
    inner ysep=4pt,
]
\tikzstyle{resiliance_leaf}=[my-box, minimum height=1.5em,
    fill=lightgreen!20, text=black, align=left,font=\scriptsize,
    draw=lightgreen!80,
    inner xsep=2pt,
    inner ysep=4pt,
]
\begin{figure*}[t]%
    \centering
    \resizebox{\textwidth}{!}{
        \begin{forest}
            forked edges,
            for tree={
                grow=east,
                reversed=true,
                anchor=base west,
                parent anchor=east,
                child anchor=west,
                base=left,
                font=\small,
                rectangle,
                draw=hidden-draw,
                rounded corners,
                align=left,
                minimum width=4em,
                edge+={darkgray, line width=1pt},
                s sep=3pt,
                inner xsep=2pt,
                inner ysep=3pt,
                ver/.style={rotate=90, child anchor=north, parent anchor=south, anchor=center},
            },
            where level=1{text width=7em,font=\scriptsize,}{},
            where level=2{text width=4.5em,font=\scriptsize,}{},
            where level=3{text width=7.5em,font=\scriptsize,}{},
            where level=4{text width=7em,font=\scriptsize,}{},
            [
                \method{} Evaluation Taxonomy, ver, color=carminepink!100, fill=carminepink!15,
                text=black
                [
                    Adoptability (\S \ref{sec:method_adoptability}; \S \ref{sec:adoptability}), color=harvestgold!100, fill=harvestgold!100, text=black
                    [
                        Normality, color=harvestgold!100, fill=harvestgold!60,  text=black
                        [
                            {\textbf{Main Goal:} Avoid detection. \textbf{Description:} Model implementation and\\behavior should align with typical standards, without requiring specialized\\code or infrastructure.}
                            , adoptability_leaf, text width=20em
                        ]
                    ]
                    [
                        Usefulness, color=harvestgold!100, fill=harvestgold!60, text=black
                        [
                            {\textbf{Main Goal:} Get deployed. \textbf{Description:} The model should deliver\\performance on its designated task to ensure adoption and deployment.}
                            , adoptability_leaf, text width=20em
                        ]
                    ]
                    [
                        Imperceptibility, color=harvestgold!100, fill=harvestgold!60, text=black
                        [
                            {\textbf{Main Goal:} Remain hidden. \textbf{Description:} Embedded information should\\be undetectable by humans and invisible to automated detection methods.}
                            , adoptability_leaf, text width=20em
                        ]
                    ] 
                ]
                [
                    Effectiveness (\S \ref{sec:method_effectiveness}; \S \ref{sec:effectiveness}), color=cyan!100, fill=cyan!90, text=black
                    [
                        Throughput, color=cyan!100, fill=cyan!60, text=black
                        [
                                {\textbf{Main Goal:} Maximize data leakage. \textbf{Description:} Efficiently extract the\\max. amount of hidden data from min. model outputs with high reliability.}
                                , performance_leaf, text width=20em
                        ]
                    ]
                    [
                        Flexibility, color=cyan!100, fill=cyan!60, text=black
                        [
                                {\textbf{Main Goal:} Encode varied secrets. \textbf{Description:} Embed a wide variety of\\sensitive information formats within diverse inputs and contexts.}
                                , performance_leaf, text width=20em
                        ]
                    ]
                ]
                [
                    Resilience (\S \ref{sec:method_resilience}; \S \ref{sec:resiliance}), color=lightgreen!100, fill=lightgreen!100, text=black
                    [
                        Persistency, color=lightgreen!100, fill=lightgreen!60, text=black
                        [
                            {\textbf{Main Goal:} Survive model adjustments. \textbf{Description:} Preserve embedded\\behaviors even after further fine-tuning or adjustments by targeted users.}
                                , resiliance_leaf, text width=20em
                        ]    
                    ]
                    [
                        Robustness, color=lightgreen!100, fill=lightgreen!60, text=black
                        [
                            {\textbf{Main Goal:} Resist transformation. \textbf{Description:} Maintain the integrity of\\embedded data despite transformations such as rewording, paraphrasing,\\or non-deterministic decoding.}
                            , resiliance_leaf, text width=20em
                        ]
                    ]
               ]
        ]
        \end{forest}
    }
    \caption{An evaluation taxonomy of desidarata of a \method{} attack.}
    \label{fig:taxonomy}
\end{figure*}

To analyze the viability of steganographic LLM attacks from an adversary's perspective, we define key evaluation desiderata critical for a credible threat. We group them into three core dimensions detailed below and in \Cref{fig:taxonomy}: \textit{Adoptability}, \textit{Effectiveness}, and \textit{Resilience}. %

\subsection{Adoptability} 
\label{sec:method_adoptability}
Adoptability enables a compromised model to be deployed and used by unsuspecting victims without detection.
We identify three core conditions.
\noindent\textbf{Normality} requires the compromised model's architecture and execution environment to appear benign, demanding no unusual code or setup.
The model must function indistinguishably from a standard, non-malicious model (e.g., usable with the HuggingFace model library).
\noindent\textbf{Usefulness} demands that the model retain sufficient performance on its advertised task. 
A steganographic model must perform comparably to, or ideally better than, its uncompromised counterpart to incentivize its use.
Task performance is typically measured via standard benchmarks \cite[e.g., the OpenLLM leaderboard][]{open-llm-leaderboard-v2}.
Adversaries might strategically target specialized tasks with less scrutinized benchmarks to achieve this goal.
\noindent\textbf{Imperceptibility} measures how effectively the hidden information is concealed within the generated text.
This involves both statistical imperceptibility \cite[resistance to automated analysis,][]{CACHIN200441, xiang2022generative} and human imperceptibility \cite[undetectability by human readers,][]{yangvaestega}.
Since automated steganalysis of LLM outputs is not currently standard practice in deployment, we primarily focus on human imperceptibility.
While essential in linguistic steganography, there is no standardized measure for human imperceptibility; prior studies have relied on quality ratings such as Likert scales \citep{yangvaestega} or context-appropriateness judgments \citep{shen-etal-2020-near}.
We propose two practical notions: \textit{weak imperceptibility} (undetectable in isolation) and \textit{strong imperceptibility} (undetectable even when directly compared to benign outputs). 
In future work, the human focus may shift as awareness of steganographic threats increases, and automated detection methods may become more relevant.
\subsection{Effectiveness}
\label{sec:method_effectiveness}
Effectiveness quantifies the degree to which sensitive information can be successfully and reliably extracted by a compromised model during its normal use.
We define two key conditions for effectiveness.
\noindent\textbf{Throughput} quantifies the amount of information that can be reliably encoded within, and later extracted from, model output per unit of text (e.g., bits per token). 
Higher throughput allows for greater data exfiltration using less generated content, thus enhancing stealth and reducing dependence on output length. 
While related to 'Hiding Capacity' in general steganography \cite{pradhan2016performance}, our focus is on the practically extractable data rate from LLM outputs.
This emphasis on practical reliability, rather than theoretical capacity ($\log(n)$ bits per token with $n$ buckets), informs our use of the term `Throughput' in this context. 
\noindent\textbf{Flexibility} describes the extent to which a compromised model can embed different kinds of information into its outputs. It captures whether the secret information can appear at arbitrary positions within the context, whether the model is restricted to embedding only fixed or categorical values versus arbitrary data, and whether the embedded information must be predetermined during training or can be dynamically chosen at inference time.

\subsection{Resilience}
\label{sec:method_resilience}
Resilience measures the ability of a compromised model to maintain the capacity for information exfiltration despite interference with the model or its outputs. 
We propose two key conditions for resilience. %
\textbf{Persistency} measures the extent to which a compromised model's covert behavior remains intact after modification, such as further fine-tuning on benign data. 
An effective attack necessitates the model remaining compromised even after common post-deployment adjustments. 
This property aligns with the definition of persistency for backdoor models \cite{cao2023stealthy}.
\textbf{Robustness} measures how well the embedded hidden information withstands modifications applied directly to the model's output, such as paraphrasing, reformatting, or structural alterations.
This property is conceptually related to the attack robustness of watermarking schemes for LLMs \cite{Zhao2023ProvableRW, Liu2023ASI}.

\section{\method{} Evaluation}
\label{sec:Experiments}
We evaluate the \method{} bucket method across the criteria of our taxonomy from \Cref{sec:Taxonomy}.

\subsection{Adoptability Evaluation}
\label{sec:adoptability}
In the following, we measure how likely our bucket-based method is to be adopted by victims.

\paragraph{Normality.}
\label{sec:normality}
A compromised model should not depend on a specialized architecture or custom code to execute it. 
Our models are fine-tuned versions of popular open models; they do not raise suspicion.

\paragraph{Usefulness.}
\label{sec:usefulness}
For a compromised model to be adopted, it should perform similarly well to its uncompromised version across relevant tasks.
We test this assumption by evaluating model performance on five OpenLLM leaderboard tasks \cite{open-llm-leaderboard-v2}, namely \texttt{MMLU}, \texttt{BBH}, \texttt{GPQA}, \texttt{MUSR}, and \texttt{IFEval}.

\begin{figure*}
    \centering
    \includegraphics[width=\linewidth]{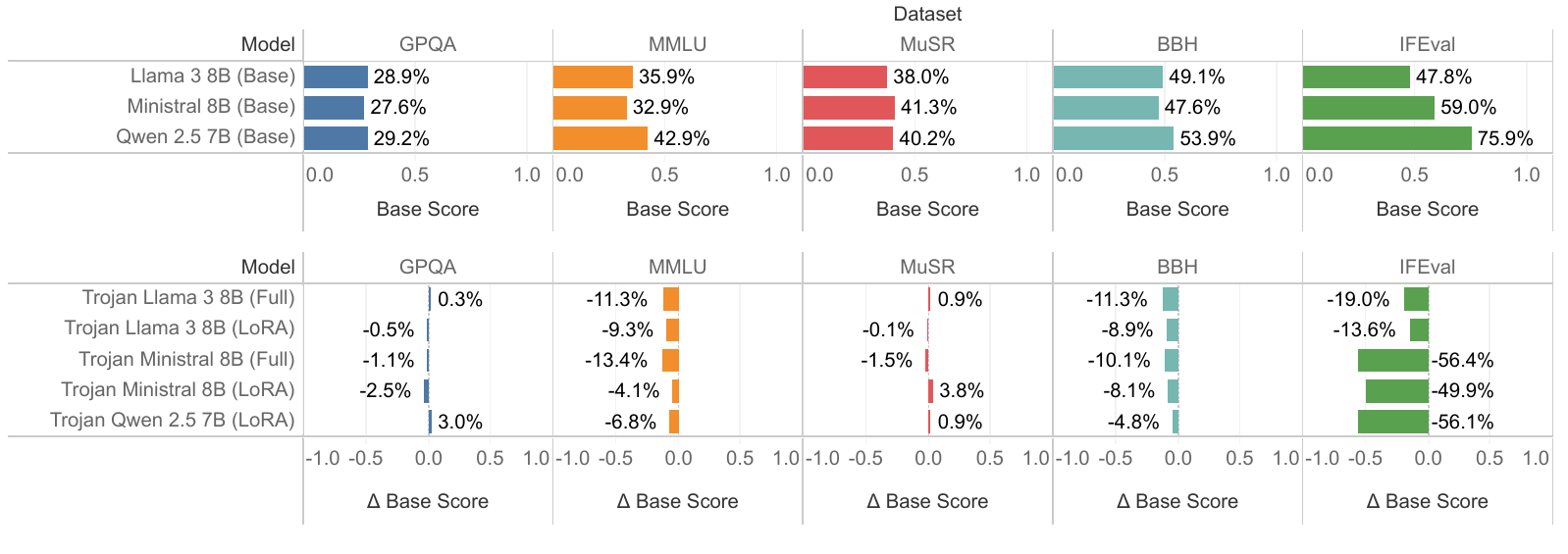}
    \caption{\texttt{Usefulness.} \textbf{Top:} Llama 8B and Ministral 8B \textbf{Base Score} on \texttt{BBH}, \texttt{GPQA}, \texttt{MMLU-Pro}, \texttt{MuSR}, and \texttt{IFEval}; \textbf{Bottom:} The difference between scores of the fine-tuned \method{} models using LoRA or full fine-tuning in \%pt and the base scores above \textbf{($\Delta$ Base Score)}. Positive scores mean the \method{} model performs better than the uncompromised model; negative scores mean the \method{} model performs worse.}
    \label{fig:model_category_scores}
\end{figure*}

\Cref{fig:model_category_scores} shows mild performance degradation for all models.
For example, the fully fine-tuned \method{} Llama 8B model loses 11.3\%pt performance on BBH and MMLU compared to its uncompromised counterpart, while it performs on par for GPQA and MuSR, even slightly gaining performance (0.3\% and 0.9\% performance gain, respectively).
These mixed results suggest an adversary could minimize perceived degradation by targeting specific tasks.
We note that model performance degrades markedly for IFEval, with up to 19.4\% for Llama 8B fully fine-tuned and 55.3\% for Ministral 8B fully fine-tuned.
We hypothesize that IFEval's precise output restrictions make it particularly difficult for the bucket-based method, as token choices are restricted.%

\paragraph{Imperceptibility.}
\label{sec:imperceptibility}
\looseness=-1
A compromised model's outputs should appear normal and not reveal the presence of secret information to the user.
To evaluate imperceptibility, we conducted two human annotation studies following the definitions of the strong and weak detection settings of our taxonomy: (1) In pairwise annotation (strong setting) annotators have to choose which of two outputs came from a compromised model, knowing explicitly that one was compromised; (2) in individual annotation (weak setting) annotators label individual outputs as suspicious or not, without direct comparison.
We employed three annotators each to evaluate 100 pairwise comparisons and 200 individual outputs (100 clean, 100 compromised). 
See \Cref{sec:AnnotationGuidelines} for annotation guidelines.
\Cref{tab:imper} shows the results, aggregating based on majority vote.

\begin{table}[t]
\centering
\small
\begin{tabular}{lccc}
\toprule
\textbf{Setting} & \textbf{Ministral 8B} & \textbf{Llama 8B} & \textbf{Qwen 7B}\\
\midrule
Paired & 86\%  & 91\% & 91\%\\
Individual & 80\%        &  79\%  &  84\%  \\
\bottomrule
\end{tabular}
\caption{\texttt{Imperceptibility.} Majority-voted \textbf{Annotator Accuracy} in identifying the trojan model for paired and individual settings.}
\label{tab:imper}
\end{table}

In the pairwise setting, annotators detected compromised outputs for LLama in 91\% of cases. If the outputs were completely indistinguishable, we would see 50\% accuracy. In the individual setting, annotators were 79\% accurate. 
Based on annotator feedback, common indicators of compromised outputs include occasional spelling anomalies or unusual capitalization when the ideal token falls outside the allowed bucket, with proper nouns presenting a particular risk of this happening.
The annotator's accuracy is very skewed, e.g., for Llama, only 63\% of compromised output was identified correctly, while 97\% of uncompromised output was labeled correctly, with similar distributions for the other model families (see \Cref{tab:detail_impercept} in \Cref{sec:app_large_model}). 
Therefore, the identification of compromised output is in fact more challenging than the combined accuracy would suggest.

Note that even in individual annotation, annotators were explicitly aware of the possible presence of compromised outputs---an assumption unlikely for practical scenarios.
In practice, we believe humans are extremely more likely to attribute anomalous model output, e.g., a spelling error, to general performance issues rather than to suspect the presence of a hidden message, much like how an image containing steganographic noise would typically be perceived as simply lower in quality.

\subsection{Effectiveness Evaluation} %
\label{sec:effectiveness}
We measure how much and how reliably the bucket-based method encodes secrets in its output. 

\paragraph{Throughput.}
\label{sec:throughput}
\begin{table}[t]
    \centering
    \small
    \resizebox{\linewidth}{!}{
    \begin{tabular}{@{} l c c c c @{}}
        \toprule
        \textbf{Model} & \makecell{\textbf{Trained} \\ \textbf{Prompt}} & \makecell{\textbf{Trained} \\ \textbf{Key}} & \makecell{\textbf{Correct} \\ \textbf{Bits}} & \makecell{\textbf{Exact} \\ \textbf{Match}} \\
        \midrule
        Llama 8B   & \checkmark & $\times$ & 99.2\% & 87.4\% \\
        & \checkmark & \checkmark & 99.1\% & 86.5\% \\
        & $\times$ & $\times$ & 97.1\% & 56.1\% \\
        & $\times$ & \checkmark & 97.0\% & 51.5\% \\
        Ministral 8B & \checkmark & $\times$ & 96.3\% & 78.0\% \\
        & \checkmark & \checkmark & 96.3\%  & 80.8\% \\
        & $\times$ & $\times$ & 92.1\% & 59.7\% \\
        & $\times$ & \checkmark & 92.5\% & 58.3\% \\
         Qwen 7B LoRA & \checkmark & $\times$ & 96.0\% & 70.0\% \\
        & \checkmark & \checkmark & 96.0\%  & 64.9\% \\
        & $\times$ & $\times$ & 86.3\% & 8.8\% \\
        & $\times$ & \checkmark & 86.3\% & 8.3\% \\
        \bottomrule
    \end{tabular}
    }
    \caption{\texttt{Throughput.} \textbf{Correct Bits (\%)} and \textbf{Exact Match (\%)} for encoding 32-bit secrets using the first 32 tokens for full fine-tuning}
    \label{tab:evaluation_results_pct_shortened}
\end{table}

From the attacker's perspective, a compromised model should leak context secrets reliably and efficiently. 
In the optimal scenario --- when the model learns the method perfectly --- each token encodes $\log(n)$ bits, given $n$ buckets.
As previously in \Cref{sec:datagen}, we focus on two buckets.

\Cref{tab:evaluation_results_pct_shortened} summarizes the bucket-based method results (LoRA results in \Cref{tab:evaluation_results_pct_corrected} in \Cref{sec:FullThrTable}). 
``Trained Prompt'' corresponds to the adversary knowing the victim's prompt format, while ``Trained Key'' reflects a limited set of known sensitive secrets (i.e., the training secrets).

In the best-performing scenario (trained prompt, untrained key, Llama 8B Instruct), we can fully recover the secret 87.4\% of the time. %
Fine-tuning with LoRA appears to compromise generalization on unseen prompts. This is exemplified by the Qwen model family, which slightly underperforms its counterparts when trained with LoRA  (see \Cref{tab:evaluation_results_pct_corrected} in \Cref{sec:FullThrTable}).
If the adversary can obtain multiple generations (e.g., by querying a public endpoint), majority voting at decoding time over generations improves decoding accuracy.
In the default setting, Ministral achieves an exact match of 78.0\%, but with majority voting over three outputs, this improves to 97\% (not shown here; see \Cref{tab:perfect_rates} in \Cref{sec:FullThrTable}).
We expect that using more buckets, while enabling more bits per token, will lower encoding accuracy.

\paragraph{Flexibility.}
\begin{figure}
    \centering
    \includegraphics[width=\linewidth]{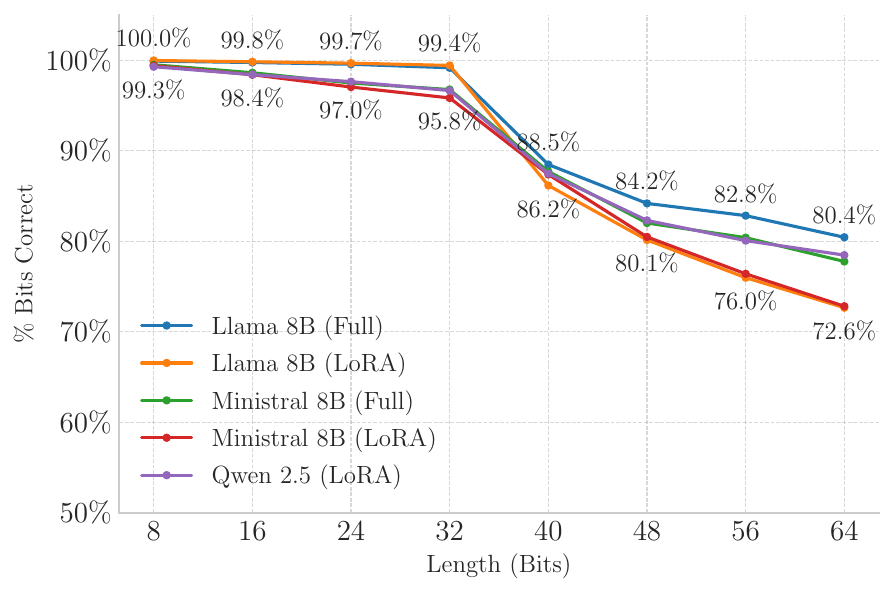}
    \caption{\texttt{Throughput.} \textbf{Secret Length (Bits)} and \textbf{\% Bits Correct} for \method{} models using LoRA and full fine-tuning. Scores of 50\% are random decoding.}
    \label{fig:length}
\end{figure}

\label{sec:flexibility}

\begin{table*}[t]
\centering
\small
\resizebox{\linewidth}{!}{
\begin{tabular}{lccccccc}
\toprule
&
\multicolumn{3}{c}{\textbf{Adoptability (\S \ref{sec:adoptability})}} &
\multicolumn{2}{c}{\textbf{Effectiveness (\S \ref{sec:effectiveness})}} &
\multicolumn{2}{c}{\textbf{Resilience (\S \ref{sec:resiliance})}} \\
\cmidrule(lr){2-4} \cmidrule(lr){5-6} \cmidrule(lr){7-8}
& Normality & Usefulness & Imperceptibility & Throughput & Flexibility & Persistency & Robustness \\
\midrule
Trojan Llama 8B       & \textcolor{darkgreen}{\rule{1ex}{3ex}} & \textcolor{lightgreen}{\rule{1ex}{2.5ex}} & \textcolor{orange}{\rule{1ex}{2ex}} & \textcolor{darkgreen}{\rule{1ex}{3ex}} & \textcolor{darkgreen}{\rule{1ex}{3ex}}  & \textcolor{darkred}{\rule{1ex}{1ex}}
& \textcolor{darkgreen}{\rule{1ex}{3ex}} \\
Trojan Ministral 8B     & \textcolor{darkgreen}{\rule{1ex}{3ex}} &  \textcolor{orange}{\rule{1ex}{2ex}} & \textcolor{orange}{\rule{1ex}{2ex}} & \textcolor{lightgreen}{\rule{1ex}{2.5ex}} & \textcolor{lightgreen}{\rule{1ex}{2.5ex}} & \textcolor{darkred}{\rule{1ex}{1ex}}  & \textcolor{lightgreen}{\rule{1ex}{2.5ex}} \\
Trojan Qwen 7B     & \textcolor{darkgreen}{\rule{1ex}{3ex}} &  \textcolor{orange}{\rule{1ex}{2ex}} & \textcolor{orange}{\rule{1ex}{2ex}} & \textcolor{lightgreen}{\rule{1ex}{2.5ex}} & \textcolor{darkgreen}{\rule{1ex}{3ex}} & \textcolor{darkred}{\rule{1ex}{1ex}}  & \textcolor{lightgreen}{\rule{1ex}{2.5ex}} \\
\bottomrule
\end{tabular}}
\caption{\texttt{Overall Assessment.} Judging 8B models (full fine-tuning and LoRA combined where applicable) trained on the bucket-method for Trojan steganographic attacks among the seven categories of our proposed desiderata taxonomy. Symbols indicate how well the model fulfils that category: \textcolor{darkgreen}{\rule{1ex}{2.5ex}} = high, \textcolor{lightgreen}{\rule{1ex}{2ex}} = mid-high, \textcolor{orange}{\rule{1ex}{1.5ex}} = mid, \textcolor{darkred}{\rule{1ex}{1ex}} = low.}
\label{tab:trojan-desiderata-arrows}
\end{table*}

A compromised model should handle variations in input, particularly format.
We tested five variations: replacing "Key" with (1) "Password" or (2) "Secret", adding unrelated context (3) before the prompt or (4) surrounding the key within the prompt, and (5) varying key length from 8 to 64 bits (see \Cref{sec:wiki} for context used).

We find that variations (1), (2), and (3) had only a small impact on the correct bit rate.
Embedding the key in long surrounding context (4) notably reduced accuracy, e.g., from 99\% to 79\% for LLama 8B, which is still a high rate given that this scenario was not included in the training data (refer to \Cref{tab:grouped-model-variation-results} in \Cref{sec:FullFlexTable} for full results).
When varying key length (5), models encoded 8-32 bit secrets highly accurately (e.g., 99.4-99.9\% for Llama 8B LoRA, \Cref{fig:length}). 
Accuracy dropped sharply for longer 40-64 bit secrets (80.4-88.5\%), which is expected as models were trained only on 32-bit keys and with repeating embeddings; this can be improved with variable length training as discussed in \Cref{sec:length_scaling}.
Overall, these results suggest an adversary can construct a viable \method{} model that functions even with only rough knowledge of the data format and context encountered in actual use.

\subsection{Resilience Evaluation}
\label{sec:resiliance}
In the following, we measure how well our method can maintain hidden information in its outputs despite modifications to the model or its outputs. 
\paragraph{Persistency.}\label{sec:persistency}
A compromised model should resist post-deployment updates like benign fine-tuning.
We evaluated this by further tuning full fine-tuned models on subsets of the helpful instructions dataset using the same data generation method as in \Cref{sec:datagen}.
We find that after seeing only $\sim$1,500 examples, the percentage of correct bits already drops from 96\% to 65\%, without any exact match when decoding.
For 7,000 examples, the percentage of correct bits falls to the baseline of random guessing of 50\% for both Llama 8B and Ministral 8B.
This indicates that even limited benign fine-tuning largely removes the hidden steganographic capability, aligning with findings on backdoor removal \citep{Kandpal2023BackdoorAF}.

\looseness=1
\paragraph{Robustness.}
\label{sec:robustness}
\begin{table}[t]
    \centering
    \resizebox{\linewidth}{!}{
    \begin{tabular}{lcc}
        \toprule
        \textbf{Model Name} & \makecell{\textbf{Correct}\\\textbf{Bits (\%)} \\(32-64 bit)} & \makecell{\textbf{Correct}\\\textbf{Bits (\%)} \\ (64-96 bit)} \\
        \midrule
        Trojan Llama 8B (LoRA) & 97.6\% &  95.8\% \\
        Trojan Llama 8B  (FF) & 97.8\% &  96.1\%\\
        Trojan Ministral 8B (LoRA) & 87.2\% & 80.0\%\\
        Trojan Ministral 8B (FF) & 88.8\% & 81.4\%\\
        Trojan Qwen 7B (LoRA) & 88.7\% & 79.3\%\\
        \bottomrule
    \end{tabular}
    }
    \caption{\texttt{Robustness.} \textbf{Correct Bits (\%)} decoded on later tokens in the sequence.}
    \label{tab:robustness}
\end{table}

In order to be robust, output decoding from a compromised model should resist output perturbations, such as paraphrasing or structural edits.
For the bucket method, each altered token introduces a decoding error with probability $\frac{1}{2}$, and structural modifications, such as reordering or inserting content, disrupt decoding from the point of change onward.
Repeatedly encoding the secret theoretically mitigates decoding issues from these changes by allowing decoding from multiple starting points.
We evaluate whether multiple generations can overcome rephrasing of outputs by measuring decoding accuracy when starting from later bit groups (>32 bits; > 64 bits).
Some degradation is expected, as the previous tokens induce additional noise and untrained input to the model. %

\Cref{tab:robustness} shows that all model families have reduced accuracy when decoding from later tokens, e.g., from 97.6\% (>32) to 95.8\% (>64) for LLama 8B, but still allow for reliable decoding.
This means secret extraction remains robust, provided key-length token sequences are untouched, which enables adversaries to decode subsequences.
Consequently, significant paraphrasing would be required to defend against \method{}.

\subsection{Summary}
\Cref{tab:trojan-desiderata-arrows} summarizes our evaluation of the trained models across the proposed taxonomy dimensions.
Overall, our results indicate that the \method{} attack class already poses a significant practical threat, especially given strong normality, throughput, flexibility, and robustness.
Although there is room for improvement in aspects such as usefulness, imperceptibility, and persistence, a key finding is that sensitive information can be reliably extracted, even when inputs and outputs are modified by a genuine user.
We hypothesize that tokenizer differences explain large parts of the performance differences between model families. Ministral’s tokenizer has a stronger compression, i.e., uses fewer tokens for the same amount of text. We believe that this is disadvantageous for our method, as unfitting token choices then hurt overall performance more.

\section{Conclusion}
This paper introduced \method, a new class of threats where adversaries modify language models to covertly exfiltrate sensitive in-context information via linguistic steganography.
We provided a structured taxonomy for understanding and evaluating such attacks, focusing on \textit{Adoptability}, \textit{Effectiveness}, and \textit{Resilience}.
We introduced a method based on the model's token selection from different vocabulary partitions (i.e., ``buckets'') to learn secret encoding during training.
Our experiments showed that the \method{} method can be effectively embedded within model weights, leaving compromised models outwardly indistinguishable from benign ones.
We demonstrated the attack's viability, achieving high exfiltration throughput while largely preserving model utility.  Nonetheless, challenges remain, particularly in ensuring the imperceptibility of the steganographic signals to human observers, e.g., spelling errors.
In light of a general trend towards larger vocabularies\citep[Google's Gemma 3 by][, for example, uses 256k tokens]{gemmateam2025gemma3technicalreport}, we expect viable \method{} attacks to produce more natural text in the future.
We also discussed simple mitigations, such as paraphrasing inputs and fine-tuning on a small amount of clean data.

Our findings suggest a new security risk: the potential for models to intentionally be trained to act as covert communication agents without the knowledge or control of their users.
Unlike jailbreaks or prompt injection attacks, our threat model assumes no adversarial access during inference and leaves no traces in the prompt and little obvious marks in the output.
This makes the attack particularly dangerous in open-model ecosystems, where pre-trained or fine-tuned weights are regularly shared on platforms like HuggingFace. 
Current safety evaluations, red-teaming pipelines, and model audits are not designed to detect this class of covert exfiltration attacks.
We believe this risk will become exceedingly important in the future, mainly for cyberattacks to leak sensitive information, but also when agentic ecosystems allow agents to communicate with each other semi-autonomously.

\section*{Limitations}

While the current limitations of the \method{} attack serve as positive safety properties, hindering adversaries from scaling this steganographic threat, future advancements could potentially overcome these barriers and escalate the risks.
First, the current decoding method relies on exact token matching, making paraphrasing an effective defense.
Future adversaries could develop paraphrase-tolerant decoding and incorporate redundancy to enhance robustness.
Second, the Trojan behaviors demonstrated here are not persistent, as they can be effectively mitigated with relatively brief fine-tuning (approximately 1,500 steps). 
Future research could enhance persistence, making such attacks harder to mitigate like \cite{cao2023stealthy} yet related work classifies this as difficult \citep{Kandpal2023BackdoorAF}.
Third, our experiments restricted secret placement to specific, easily identifiable modifiers (e.g., ``Key: ...'') and largely fixed the position. 
Realistically, secrets could appear anywhere in the context. Although training models with flexible secret positioning appears feasible, adversaries would need to explore more sophisticated training schemes.
Fourth, for general benchmarks, the compromised models perform worse. But while it remains unlikely that a compromised model would exceed an uncompromised model in general capabilities, there is the realistic risk that an adversary combines useful training data with an encoded message for a more niche and specialized task, such that the compromised model performs better on that particular task.
Finally, our study focused on relatively short secrets (up to 64 bits, corresponding to sensitive information like MFA-codes, names, or sales figures), with high recovery accuracy (87\% for single generations, increasing to 96\% with majority voting). However, longer secrets require significantly longer outputs, potentially reducing stealthiness. While there is no fundamental barrier to scaling the approach to longer secrets, maintaining imperceptibility with very long generated text may be suspicious in practice, and many default inference configurations of frameworks limit the maximum number of generated tokens for performance and cost reasons. Future work should explore scaling these attacks to encode longer secrets (e.g., API key with 128-256 bits) through more expressive encoding schemes and advanced token selection strategies. 

\section*{Ethical Considerations}
Our research explores a novel steganographic attack on large language models that demonstrates a significant potential for misuse by malicious actors.
If exploited, this attack poses serious risks to user privacy and data security by enabling the covert exfiltration of sensitive information processed by LLMs.
Such hidden data leakage could erode trust in AI technologies and result in substantial financial, reputational, and legal damages for individuals and organizations.

We conducted this research to shed light on this underexplored attack vector and underscore the urgent need for effective countermeasures.
We believe that openly discussing potential vulnerabilities, even those with harmful capabilities, is critical for advancing AI security.
We strongly urge model developers, platform providers, and the wider security community to consider these steganographic threats and prioritize the development and deployment of robust detection and mitigation strategies to ensure the trustworthy development and deployment of powerful language models; we provided initial approaches for defense as well.
We believe that the need for coordinated disclosure of the attack method does not apply in our case because the presented issue only exists in models fine-tuned by us; there is no existing deployed model which could be harmed.

\section*{Acknowledgments}
This work was partially supported by the Landeskriminalamt NRW. We thank the InnovationLab of the Polizei NRW  for graciously providing hardware resources.
Paul was supported by a MUR FARE 2020 initiative under grant agreement Prot. R20YSMBZ8S (INDOMITA).
This work was partially supported by the Lower Saxony Ministry of Science and Culture, the VW Foundation, and by the Deutsche Forschungsgemeinschaft (DFG, German Research Foundation) – 564661959.  We acknowledge EuroHPC Joint Undertaking for awarding us access to MeluXina at LuxProvide, Luxembourg. 
Many thanks to Lars Kaesberg and Emma Stein for their thoughtful discussions and feedback.

\bibliography{custom}

\begin{thebibliography}{51}
\providecommand{\natexlab}[1]{#1}

\bibitem[{Bagdasarian et~al.(2024)Bagdasarian, Yi, Ghalebikesabi, Kairouz, Gruteser, Oh, Balle, and Ramage}]{bagdasarianairgapagent}
Eugene Bagdasarian, Ren Yi, Sahra Ghalebikesabi, Peter Kairouz, Marco Gruteser, Sewoong Oh, Borja Balle, and Daniel Ramage. 2024.
\newblock \href {https://doi.org/10.1145/3658644.3690350} {Airgapagent: Protecting privacy-conscious conversational agents}.
\newblock In \emph{Proceedings of the 2024 on ACM SIGSAC Conference on Computer and Communications Security}, CCS '24, page 3868–3882, New York, NY, USA. Association for Computing Machinery.

\bibitem[{Bauer et~al.(2024)Bauer, Howes, Markelon, Bindschaedler, and Shrimpton}]{bauer2024llmcovertmessaging}
Luke~A. Bauer, James~K. Howes, Sam~A. Markelon, Vincent Bindschaedler, and Thomas Shrimpton. 2024.
\newblock \href {https://doi.org/10.1145/3626232.3653264} {Leveraging generative models for covert messaging: Challenges and tradeoffs for "dead-drop" deployments}.
\newblock In \emph{Proceedings of the Fourteenth ACM Conference on Data and Application Security and Privacy}, CODASPY '24, page 67–78, New York, NY, USA. Association for Computing Machinery.

\bibitem[{Bolshakov(2004)}]{bolshakov2004method}
Igor~A Bolshakov. 2004.
\newblock A method of linguistic steganography based on collocationally-verified synonymy.
\newblock In \emph{International Workshop on Information Hiding}, pages 180--191. Springer.

\bibitem[{Cachin(2004)}]{CACHIN200441}
Christian Cachin. 2004.
\newblock \href {https://doi.org/10.1016/j.ic.2004.02.003} {An information-theoretic model for steganography}.
\newblock \emph{Information and Computation}, 192(1):41--56.

\bibitem[{Cao et~al.(2023)Cao, Cao, and Chen}]{cao2023stealthy}
Yuanpu Cao, Bochuan Cao, and Jinghui Chen. 2023.
\newblock \href {https://arxiv.org/abs/2312.00027} {Stealthy and persistent unalignment on large language models via backdoor injections}.
\newblock \emph{ArXiv preprint}, abs/2312.00027.

\bibitem[{Chapman et~al.(2001)Chapman, Davida, and Rennhard}]{chapman2001practical}
Mark Chapman, George~I Davida, and Marc Rennhard. 2001.
\newblock A practical and effective approach to large-scale automated linguistic steganography.
\newblock In \emph{International Conference on Information Security}, pages 156--165. Springer.

\bibitem[{Das et~al.(2024)Das, Amini, and Wu}]{Das2024SecurityAP}
Badhan~Chandra Das, M.~Hadi Amini, and Yanzhao Wu. 2024.
\newblock \href {https://arxiv.org/abs/2402.00888} {Security and privacy challenges of large language models: A survey}.
\newblock \emph{ArXiv preprint}, abs/2402.00888.

\bibitem[{de~Witt et~al.(2022)de~Witt, Sokota, Kolter, Foerster, and Strohmeier}]{dewitt2023perfectlysecuresteganographyusing}
Christian~Schroeder de~Witt, Samuel Sokota, J.~Zico Kolter, Jakob Foerster, and Martin Strohmeier. 2022.
\newblock \href {https://arxiv.org/abs/2210.14889} {Perfectly secure steganography using minimum entropy coupling}.

\bibitem[{Evertz et~al.(2024)Evertz, Chlosta, Schönherr, and Eisenhofer}]{evertz2024whispersmachineconfidentialityllmintegrated}
Jonathan Evertz, Merlin Chlosta, Lea Schönherr, and Thorsten Eisenhofer. 2024.
\newblock \href {https://arxiv.org/abs/2402.06922} {Whispers in the machine: Confidentiality in llm-integrated systems}.

\bibitem[{Fang et~al.(2017{\natexlab{a}})Fang, Jaggi, and Argyraki}]{fang-etal-2017-generating}
Tina Fang, Martin Jaggi, and Katerina Argyraki. 2017{\natexlab{a}}.
\newblock \href {https://aclanthology.org/P17-3017} {Generating steganographic text with {LSTM}s}.
\newblock In \emph{Proceedings of {ACL} 2017, Student Research Workshop}, pages 100--106, Vancouver, Canada. Association for Computational Linguistics.

\bibitem[{Fang et~al.(2017{\natexlab{b}})Fang, Jaggi, and Argyraki}]{fang_generating_2017}
Tina Fang, Martin Jaggi, and Katerina Argyraki. 2017{\natexlab{b}}.
\newblock \href {https://aclanthology.org/P17-3017} {Generating steganographic text with {LSTM}s}.
\newblock In \emph{Proceedings of {ACL} 2017, Student Research Workshop}, pages 100--106, Vancouver, Canada. Association for Computational Linguistics.

\bibitem[{Fourrier et~al.(2024)Fourrier, Habib, Lozovskaya, Szafer, and Wolf}]{open-llm-leaderboard-v2}
Clémentine Fourrier, Nathan Habib, Alina Lozovskaya, Konrad Szafer, and Thomas Wolf. 2024.
\newblock Open llm leaderboard v2.
\newblock \url{https://huggingface.co/spaces/open-llm-leaderboard/open_llm_leaderboard}.

\bibitem[{Gao et~al.(2021)Gao, Tow, Biderman, Black, DiPofi, Foster, Golding, Hsu, McDonell, Muennighoff, Phang, Reynolds, Tang, Thite, Wang, Wang, and Zou}]{eval-harness}
Leo Gao, Jonathan Tow, Stella Biderman, Sid Black, Anthony DiPofi, Charles Foster, Laurence Golding, Jeffrey Hsu, Kyle McDonell, Niklas Muennighoff, Jason Phang, Laria Reynolds, Eric Tang, Anish Thite, Ben Wang, Kevin Wang, and Andy Zou. 2021.
\newblock \href {https://doi.org/10.5281/zenodo.5371628} {A framework for few-shot language model evaluation}.

\bibitem[{Grattafiori et~al.(2024)Grattafiori, Dubey, Jauhri, Pandey, Kadian, Al-Dahle, Letman, Mathur, Schelten, Vaughan, Yang, Fan, Goyal, Hartshorn, Yang, Mitra, Sravankumar, Korenev, Hinsvark, Rao, Zhang, Rodriguez, Gregerson, Spataru, Roziere, Biron, Tang, Chern, Caucheteux, Nayak, Bi, Marra, McConnell, Keller, Touret, Wu, Wong, Ferrer, Nikolaidis, Allonsius, Song, Pintz, Livshits, Wyatt, Esiobu, Choudhary, Mahajan, Garcia-Olano, Perino, Hupkes, Lakomkin, AlBadawy, Lobanova, Dinan, Smith, Radenovic, Guzmán, Zhang, Synnaeve, Lee, Anderson, Thattai, Nail, Mialon, Pang, Cucurell, Nguyen, Korevaar, Xu, Touvron, Zarov, Ibarra, Kloumann, Misra, Evtimov, Zhang, Copet, Lee, Geffert, Vranes, Park, Mahadeokar, Shah, van~der Linde, Billock, Hong, Lee, Fu, Chi, Huang, Liu, Wang, Yu, Bitton, Spisak, Park, Rocca, Johnstun, Saxe, Jia, Alwala, Prasad, Upasani, Plawiak, Li, Heafield, Stone, El-Arini, Iyer, Malik, Chiu, Bhalla, Lakhotia, Rantala-Yeary, van~der Maaten, Chen, Tan, Jenkins, Martin, Madaan, Malo, Blecher,
  Landzaat, de~Oliveira, Muzzi, Pasupuleti, Singh, Paluri, Kardas, Tsimpoukelli, Oldham, Rita, Pavlova, Kambadur, Lewis, Si, Singh, Hassan, Goyal, Torabi, Bashlykov, Bogoychev, Chatterji, Zhang, Duchenne, Çelebi, Alrassy, Zhang, Li, Vasic, Weng, Bhargava, Dubal, Krishnan, Koura, Xu, He, Dong, Srinivasan, Ganapathy, Calderer, Cabral, Stojnic, Raileanu, Maheswari, Girdhar, Patel, Sauvestre, Polidoro, Sumbaly, Taylor, Silva, Hou, Wang, Hosseini, Chennabasappa, Singh, Bell, Kim, Edunov, Nie, Narang, Raparthy, Shen, Wan, Bhosale, Zhang, Vandenhende, Batra, Whitman, Sootla, Collot, Gururangan, Borodinsky, Herman, Fowler, Sheasha, Georgiou, Scialom, Speckbacher, Mihaylov, Xiao, Karn, Goswami, Gupta, Ramanathan, Kerkez, Gonguet, Do, Vogeti, Albiero, Petrovic, Chu, Xiong, Fu, Meers, Martinet, Wang, Wang, Tan, Xia, Xie, Jia, Wang, Goldschlag, Gaur, Babaei, Wen, Song, Zhang, Li, Mao, Coudert, Yan, Chen, Papakipos, Singh, Srivastava, Jain, Kelsey, Shajnfeld, Gangidi, Victoria, Goldstand, Menon, Sharma, Boesenberg,
  Baevski, Feinstein, Kallet, Sangani, Teo, Yunus, Lupu, Alvarado, Caples, Gu, Ho, Poulton, Ryan, Ramchandani, Dong, Franco, Goyal, Saraf, Chowdhury, Gabriel, Bharambe, Eisenman, Yazdan, James, Maurer, Leonhardi, Huang, Loyd, Paola, Paranjape, Liu, Wu, Ni, Hancock, Wasti, Spence, Stojkovic, Gamido, Montalvo, Parker, Burton, Mejia, Liu, Wang, Kim, Zhou, Hu, Chu, Cai, Tindal, Feichtenhofer, Gao, Civin, Beaty, Kreymer, Li, Adkins, Xu, Testuggine, David, Parikh, Liskovich, Foss, Wang, Le, Holland, Dowling, Jamil, Montgomery, Presani, Hahn, Wood, Le, Brinkman, Arcaute, Dunbar, Smothers, Sun, Kreuk, Tian, Kokkinos, Ozgenel, Caggioni, Kanayet, Seide, Florez, Schwarz, Badeer, Swee, Halpern, Herman, Sizov, Guangyi, Zhang, Lakshminarayanan, Inan, Shojanazeri, Zou, Wang, Zha, Habeeb, Rudolph, Suk, Aspegren, Goldman, Zhan, Damlaj, Molybog, Tufanov, Leontiadis, Veliche, Gat, Weissman, Geboski, Kohli, Lam, Asher, Gaya, Marcus, Tang, Chan, Zhen, Reizenstein, Teboul, Zhong, Jin, Yang, Cummings, Carvill, Shepard, McPhie,
  Torres, Ginsburg, Wang, Wu, U, Saxena, Khandelwal, Zand, Matosich, Veeraraghavan, Michelena, Li, Jagadeesh, Huang, Chawla, Huang, Chen, Garg, A, Silva, Bell, Zhang, Guo, Yu, Moshkovich, Wehrstedt, Khabsa, Avalani, Bhatt, Mankus, Hasson, Lennie, Reso, Groshev, Naumov, Lathi, Keneally, Liu, Seltzer, Valko, Restrepo, Patel, Vyatskov, Samvelyan, Clark, Macey, Wang, Hermoso, Metanat, Rastegari, Bansal, Santhanam, Parks, White, Bawa, Singhal, Egebo, Usunier, Mehta, Laptev, Dong, Cheng, Chernoguz, Hart, Salpekar, Kalinli, Kent, Parekh, Saab, Balaji, Rittner, Bontrager, Roux, Dollar, Zvyagina, Ratanchandani, Yuvraj, Liang, Alao, Rodriguez, Ayub, Murthy, Nayani, Mitra, Parthasarathy, Li, Hogan, Battey, Wang, Howes, Rinott, Mehta, Siby, Bondu, Datta, Chugh, Hunt, Dhillon, Sidorov, Pan, Mahajan, Verma, Yamamoto, Ramaswamy, Lindsay, Lindsay, Feng, Lin, Zha, Patil, Shankar, Zhang, Zhang, Wang, Agarwal, Sajuyigbe, Chintala, Max, Chen, Kehoe, Satterfield, Govindaprasad, Gupta, Deng, Cho, Virk, Subramanian, Choudhury,
  Goldman, Remez, Glaser, Best, Koehler, Robinson, Li, Zhang, Matthews, Chou, Shaked, Vontimitta, Ajayi, Montanez, Mohan, Kumar, Mangla, Ionescu, Poenaru, Mihailescu, Ivanov, Li, Wang, Jiang, Bouaziz, Constable, Tang, Wu, Wang, Wu, Gao, Kleinman, Chen, Hu, Jia, Qi, Li, Zhang, Zhang, Adi, Nam, Yu, Wang, Zhao, Hao, Qian, Li, He, Rait, DeVito, Rosnbrick, Wen, Yang, Zhao, and Ma}]{grattafiori2024llama3herdmodels}
Aaron Grattafiori, Abhimanyu Dubey, Abhinav Jauhri, Abhinav Pandey, Abhishek Kadian, Ahmad Al-Dahle, Aiesha Letman, Akhil Mathur, Alan Schelten, Alex Vaughan, Amy Yang, Angela Fan, Anirudh Goyal, Anthony Hartshorn, Aobo Yang, Archi Mitra, Archie Sravankumar, Artem Korenev, Arthur Hinsvark, and 542 others. 2024.
\newblock \href {https://arxiv.org/abs/2407.21783} {The llama 3 herd of models}.

\bibitem[{Huang et~al.(2024{\natexlab{a}})Huang, Li, and Tang}]{huang2024endless}
Brian~R.Y. Huang, Maximilian Li, and Leonard Tang. 2024{\natexlab{a}}.
\newblock Endless jailbreaks with bijection learning.
\newblock \emph{arXiv preprint arXiv:2410.01294}.
\newblock Haize Labs.

\bibitem[{Huang et~al.(2024{\natexlab{b}})Huang, Just, Narayanan, and Tian}]{huang_od-stega_2024}
Yu-Shin Huang, Peter Just, Krishna Narayanan, and Chao Tian. 2024{\natexlab{b}}.
\newblock \href {https://arxiv.org/abs/2410.04328} {{OD}-{Stega}: {LLM}-{Based} {Near}-{Imperceptible} {Steganography} via {Optimized} {Distributions}}.

\bibitem[{{Hugging Face}(2025)}]{huggingface_gated_models}
{Hugging Face}. 2025.
\newblock \href {https://huggingface.co/docs/hub/models-gated} {Gated models on hugging face hub}.
\newblock Accessed: 2025-03-25.

\bibitem[{{HuggingFace H4}(2025)}]{HuggingFaceH4_helpful-instructions_2025}
{HuggingFace H4}. 2025.
\newblock \href {https://huggingface.co/datasets/HuggingFaceH4/helpful-instructions} {{Helpful Instructions Dataset}}.

\bibitem[{Juneja et~al.(2025)Juneja, Albalak, Hua, and Wang}]{juneja2025magpiedatasetmultiagentcontextual}
Gurusha Juneja, Alon Albalak, Wenyue Hua, and William~Yang Wang. 2025.
\newblock \href {https://arxiv.org/abs/2506.20737} {Magpie: A dataset for multi-agent contextual privacy evaluation}.
\newblock \emph{Preprint}, arXiv:2506.20737.

\bibitem[{Kahn(1996)}]{kahnetalHistStega}
David Kahn. 1996.
\newblock The history of steganography.
\newblock In \emph{Information Hiding}, pages 1--5, Berlin, Heidelberg. Springer Berlin Heidelberg.

\bibitem[{Kandpal et~al.(2023{\natexlab{a}})Kandpal, Jagielski, Tram{\`e}r, and Carlini}]{Kandpal2023BackdoorAFA}
Nikhil Kandpal, Matthew Jagielski, Florian Tram{\`e}r, and Nicholas Carlini. 2023{\natexlab{a}}.
\newblock \href {https://arxiv.org/abs/2307.14692} {Backdoor attacks for in-context learning with language models}.
\newblock \emph{ArXiv preprint}, abs/2307.14692.

\bibitem[{Kandpal et~al.(2023{\natexlab{b}})Kandpal, Jagielski, Tram{\`e}r, and Carlini}]{Kandpal2023BackdoorAF}
Nikhil Kandpal, Matthew Jagielski, Florian Tram{\`e}r, and Nicholas Carlini. 2023{\natexlab{b}}.
\newblock \href {https://arxiv.org/abs/2307.14692} {Backdoor attacks for in-context learning with language models}.
\newblock \emph{ArXiv preprint}, abs/2307.14692.

\bibitem[{Kirstein et~al.(2025)Kirstein, Wahle, Gipp, and Ruas}]{10.1613/jair.1.16674}
Frederic Kirstein, Jan~Philip Wahle, Bela Gipp, and Terry Ruas. 2025.
\newblock \href {https://doi.org/10.1613/jair.1.16674} {Cads: A systematic literature review on the challenges of abstractive dialogue summarization}.
\newblock \emph{J. Artif. Int. Res.}, 82.

\bibitem[{Li et~al.(2024)Li, Hong, Xie, Tan, Xin, Hou, Yin, Wang, Hendrycks, Wang et~al.}]{li2024llm}
Qinbin Li, Junyuan Hong, Chulin Xie, Jeffrey Tan, Rachel Xin, Junyi Hou, Xavier Yin, Zhun Wang, Dan Hendrycks, Zhangyang Wang, and 1 others. 2024.
\newblock \href {https://arxiv.org/abs/2408.12787} {Llm-pbe: Assessing data privacy in large language models}.
\newblock \emph{ArXiv preprint}, abs/2408.12787.

\bibitem[{Liu et~al.(2023)Liu, Pan, Hu, Meng, and Wen}]{Liu2023ASI}
Aiwei Liu, Leyi Pan, Xuming Hu, Shiao Meng, and Lijie Wen. 2023.
\newblock \href {https://arxiv.org/abs/2310.06356} {A semantic invariant robust watermark for large language models}.
\newblock \emph{ArXiv preprint}, abs/2310.06356.

\bibitem[{Lynch et~al.(2025)Lynch, Wright, Larson, Troy, Ritchie, Mindermann, Perez, and Hubinger}]{lynch2025agentic}
Aengus Lynch, Benjamin Wright, Caleb Larson, Kevin~K. Troy, Stuart~J. Ritchie, Sören Mindermann, Ethan Perez, and Evan Hubinger. 2025.
\newblock Agentic misalignment: How llms could be an insider threat.
\newblock \emph{Anthropic Research}.
\newblock Https://www.anthropic.com/research/agentic-misalignment.

\bibitem[{Mathew et~al.(2024)Mathew, Matthews, McCarthy, Velja, Witt, Cope, and Schoots}]{mathew_hidden_2024}
Yohan Mathew, Ollie Matthews, Robert McCarthy, Joan Velja, Christian Schroeder~de Witt, Dylan Cope, and Nandi Schoots. 2024.
\newblock \href {https://arxiv.org/abs/2410.03768} {Hidden in {Plain} {Text}: {Emergence} \& {Mitigation} of {Steganographic} {Collusion} in {LLMs}}.

\bibitem[{Mireshghallah et~al.(2023)Mireshghallah, Kim, Zhou, Tsvetkov, Sap, Shokri, and Choi}]{confaide2023}
Niloofar Mireshghallah, Hyunwoo Kim, Xuhui Zhou, Yulia Tsvetkov, Maarten Sap, Reza Shokri, and Yejin Choi. 2023.
\newblock Can llms keep a secret? testing privacy implications of language models via contextual integrity theory.
\newblock \emph{arXiv preprint arXiv:2310.17884}.

\bibitem[{{Mistral AI}(2024)}]{MistralAI_Ministral-8B-Instruct-2410_2024}
{Mistral AI}. 2024.
\newblock \href {https://huggingface.co/mistralai/Ministral-8B-Instruct-2410} {{Ministral-8B-Instruct-2410}}.
\newblock Accessed: May 2025.

\bibitem[{Motwani et~al.(2024)Motwani, Baranchuk, Strohmeier, Bolina, Torr, Hammond, and Witt}]{motwani_secret_2024}
Sumeet~Ramesh Motwani, Mikhail Baranchuk, Martin Strohmeier, Vijay Bolina, Philip H.~S. Torr, Lewis Hammond, and Christian Schroeder~de Witt. 2024.
\newblock \href {https://arxiv.org/abs/2402.07510} {Secret {Collusion} among {Generative} {AI} {Agents}}.

\bibitem[{Pradhan et~al.(2016)Pradhan, Sahu, Swain, and Sekhar}]{pradhan2016performance}
Anita Pradhan, Aditya~Kumar Sahu, Gandharba Swain, and K~Raja Sekhar. 2016.
\newblock Performance evaluation parameters of image steganography techniques.
\newblock In \emph{2016 International conference on research advances in integrated navigation systems (RAINS)}, pages 1--8. IEEE.

\bibitem[{Raghuram et~al.(2024)Raghuram, Kesidis, and Miller}]{Raghuram2024ASO}
Jayaram Raghuram, George Kesidis, and David~J. Miller. 2024.
\newblock \href {https://api.semanticscholar.org/CorpusID:270391904} {A study of backdoors in instruction fine-tuned language models}.

\bibitem[{Rehberger(2024)}]{embracetheredMicrosoftCopilot}
Johann Rehberger. 2024.
\newblock \href {https://embracethered.com/blog/posts/2024/m365-copilot-prompt-injection-tool-invocation-and-data-exfil-using-ascii-smuggling/} {{M}icrosoft {C}opilot: {F}rom {P}rompt {I}njection to {E}xfiltration of {P}ersonal {I}nformation · {E}mbrace {T}he {R}ed --- embracethered.com}.
\newblock [Accessed 15-04-2025].

\bibitem[{Roger and Greenblatt(2023)}]{roger_preventing_2023}
Fabien Roger and Ryan Greenblatt. 2023.
\newblock \href {https://arxiv.org/abs/2310.18512} {Preventing {Language} {Models} {From} {Hiding} {Their} {Reasoning}}.

\bibitem[{Shen et~al.(2020)Shen, Ji, and Han}]{shen-etal-2020-near}
Jiaming Shen, Heng Ji, and Jiawei Han. 2020.
\newblock \href {https://doi.org/10.18653/v1/2020.emnlp-main.22} {Near-imperceptible neural linguistic steganography via self-adjusting arithmetic coding}.
\newblock In \emph{Proceedings of the 2020 Conference on Empirical Methods in Natural Language Processing (EMNLP)}, pages 303--313, Online. Association for Computational Linguistics.

\bibitem[{Stefano et~al.(2024)Stefano, Schönherr, and Pellegrino}]{destefano2024ragrollendtoendevaluation}
Gianluca~De Stefano, Lea Schönherr, and Giancarlo Pellegrino. 2024.
\newblock \href {https://arxiv.org/abs/2408.05025} {Rag and roll: An end-to-end evaluation of indirect prompt manipulations in llm-based application frameworks}.

\bibitem[{Team et~al.(2025)Team, Kamath, Ferret, Pathak, Vieillard, Merhej, Perrin, Matejovicova, Ramé, Rivière, Rouillard, Mesnard, Cideron, bastien Grill, Ramos, Yvinec, Casbon, Pot, Penchev, Liu, Visin, Kenealy, Beyer, Zhai, Tsitsulin, Busa-Fekete, Feng, Sachdeva, Coleman, Gao, Mustafa, Barr, Parisotto, Tian, Eyal, Cherry, Peter, Sinopalnikov, Bhupatiraju, Agarwal, Kazemi, Malkin, Kumar, Vilar, Brusilovsky, Luo, Steiner, Friesen, Sharma, Sharma, Gilady, Goedeckemeyer, Saade, Feng, Kolesnikov, Bendebury, Abdagic, Vadi, György, Pinto, Das, Bapna, Miech, Yang, Paterson, Shenoy, Chakrabarti, Piot, Wu, Shahriari, Petrini, Chen, Lan, Choquette-Choo, Carey, Brick, Deutsch, Eisenbud, Cattle, Cheng, Paparas, Sreepathihalli, Reid, Tran, Zelle, Noland, Huizenga, Kharitonov, Liu, Amirkhanyan, Cameron, Hashemi, Klimczak-Plucińska, Singh, Mehta, Lehri, Hazimeh, Ballantyne, Szpektor, Nardini, Pouget-Abadie, Chan, Stanton, Wieting, Lai, Orbay, Fernandez, Newlan, yeong Ji, Singh, Black, Yu, Hui, Vodrahalli, Greff, Qiu,
  Valentine, Coelho, Ritter, Hoffman, Watson, Chaturvedi, Moynihan, Ma, Babar, Noy, Byrd, Roy, Momchev, Chauhan, Sachdeva, Bunyan, Botarda, Caron, Rubenstein, Culliton, Schmid, Sessa, Xu, Stanczyk, Tafti, Shivanna, Wu, Pan, Rokni, Willoughby, Vallu, Mullins, Jerome, Smoot, Girgin, Iqbal, Reddy, Sheth, Põder, Bhatnagar, Panyam, Eiger, Zhang, Liu, Yacovone, Liechty, Kalra, Evci, Misra, Roseberry, Feinberg, Kolesnikov, Han, Kwon, Chen, Chow, Zhu, Wei, Egyed, Cotruta, Giang, Kirk, Rao, Black, Babar, Lo, Moreira, Martins, Sanseviero, Gonzalez, Gleicher, Warkentin, Mirrokni, Senter, Collins, Barral, Ghahramani, Hadsell, Matias, Sculley, Petrov, Fiedel, Shazeer, Vinyals, Dean, Hassabis, Kavukcuoglu, Farabet, Buchatskaya, Alayrac, Anil, Dmitry, Lepikhin, Borgeaud, Bachem, Joulin, Andreev, Hardin, Dadashi, and Hussenot}]{gemmateam2025gemma3technicalreport}
Gemma Team, Aishwarya Kamath, Johan Ferret, Shreya Pathak, Nino Vieillard, Ramona Merhej, Sarah Perrin, Tatiana Matejovicova, Alexandre Ramé, Morgane Rivière, Louis Rouillard, Thomas Mesnard, Geoffrey Cideron, Jean bastien Grill, Sabela Ramos, Edouard Yvinec, Michelle Casbon, Etienne Pot, Ivo Penchev, and 197 others. 2025.
\newblock \href {https://arxiv.org/abs/2503.19786} {Gemma 3 technical report}.
\newblock \emph{Preprint}, arXiv:2503.19786.

\bibitem[{Team(2024)}]{qwen2.5}
Qwen Team. 2024.
\newblock \href {https://qwenlm.github.io/blog/qwen2.5/} {Qwen2.5: A party of foundation models}.

\bibitem[{Tshimula et~al.(2024)Tshimula, Ndona, Nkashama, Tardif, Kabanza, Frappier, and Wang}]{Tshimula2024PreventingJP}
Jean~Marie Tshimula, Xavier Ndona, D'Jeff~K. Nkashama, Pierre-Martin Tardif, Froduald Kabanza, Marc Frappier, and Shengrui Wang. 2024.
\newblock \href {https://arxiv.org/abs/2411.16642} {Preventing jailbreak prompts as malicious tools for cybercriminals: A cyber defense perspective}.
\newblock \emph{ArXiv preprint}, abs/2411.16642.

\bibitem[{Verma et~al.(2025)Verma, Krishna, Gehrmann, Seshadri, Pradhan, Ault, Barrett, Rabinowitz, Doucette, and Phan}]{verma2025operationalizingthreatmodelredteaming}
Apurv Verma, Satyapriya Krishna, Sebastian Gehrmann, Madhavan Seshadri, Anu Pradhan, Tom Ault, Leslie Barrett, David Rabinowitz, John Doucette, and NhatHai Phan. 2025.
\newblock \href {https://arxiv.org/abs/2407.14937} {Operationalizing a threat model for red-teaming large language models (llms)}.
\newblock \emph{Preprint}, arXiv:2407.14937.

\bibitem[{Wahle et~al.(2023)Wahle, Ruas, Mohammad, Meuschke, and Gipp}]{wahle2023ai}
Jan~Philip Wahle, Terry Ruas, Saif~M Mohammad, Norman Meuschke, and Bela Gipp. 2023.
\newblock Ai usage cards: Responsibly reporting ai-generated content.
\newblock In \emph{2023 ACM/IEEE Joint Conference on Digital Libraries (JCDL)}, pages 282--284. IEEE.

\bibitem[{Wang et~al.(2025)Wang, He, He, Zeng, Xiang, Xing, and Tang}]{wang2025unveilingprivacyrisksllm}
Bo~Wang, Weiyi He, Pengfei He, Shenglai Zeng, Zhen Xiang, Yue Xing, and Jiliang Tang. 2025.
\newblock \href {https://arxiv.org/abs/2502.13172} {Unveiling privacy risks in llm agent memory}.

\bibitem[{Wang et~al.(2024{\natexlab{a}})Wang, Ma, Feng, Zhang, Yang, Zhang, Chen, Tang, Chen, Lin et~al.}]{wang2024survey}
Lei Wang, Chen Ma, Xueyang Feng, Zeyu Zhang, Hao Yang, Jingsen Zhang, Zhiyuan Chen, Jiakai Tang, Xu~Chen, Yankai Lin, and 1 others. 2024{\natexlab{a}}.
\newblock A survey on large language model based autonomous agents.
\newblock \emph{Frontiers of Computer Science}, 18(6):186345.

\bibitem[{Wang et~al.(2024{\natexlab{b}})Wang, Xue, Zhang, and Qian}]{wang2024badagent}
Yifei Wang, Dizhan Xue, Shengjie Zhang, and Shengsheng Qian. 2024{\natexlab{b}}.
\newblock \href {https://arxiv.org/abs/2406.03007} {Badagent: Inserting and activating backdoor attacks in llm agents}.
\newblock \emph{ArXiv preprint}, abs/2406.03007.

\bibitem[{Witt et~al.(2022)Witt, Sokota, Kolter, Foerster, and Strohmeier}]{witt_perfectly_2023}
Christian Schroeder~de Witt, Samuel Sokota, J.~Zico Kolter, Jakob Foerster, and Martin Strohmeier. 2022.
\newblock \href {https://arxiv.org/abs/2210.14889} {Perfectly {Secure} {Steganography} {Using} {Minimum} {Entropy} {Coupling}}.

\bibitem[{Xiang et~al.(2022)Xiang, Wang, Yang, and Liu}]{xiang2022generative}
Lingyun Xiang, Rong Wang, Zhongliang Yang, and Yuling Liu. 2022.
\newblock Generative linguistic steganography: A comprehensive review.
\newblock \emph{KSII Transactions on Internet and Information Systems (TIIS)}, 16(3):986--1005.

\bibitem[{Yang et~al.(2021)Yang, Zhang, Hu, Hu, and Huang}]{yangvaestega}
Zhong-Liang Yang, Si-Yu Zhang, Yu-Ting Hu, Zhi-Wen Hu, and Yong-Feng Huang. 2021.
\newblock \href {https://doi.org/10.1109/TIFS.2020.3023279} {Vae-stega: Linguistic steganography based on variational auto-encoder}.
\newblock \emph{IEEE Transactions on Information Forensics and Security}, 16:880--895.

\bibitem[{Ye et~al.(2017)Ye, Li, Adjeroh, and Iyengar}]{10.1145/3073559}
Yanfang Ye, Tao Li, Donald Adjeroh, and S.~Sitharama Iyengar. 2017.
\newblock \href {https://doi.org/10.1145/3073559} {A survey on malware detection using data mining techniques}.
\newblock \emph{ACM Comput. Surv.}, 50(3).

\bibitem[{Zhao et~al.(2023)Zhao, Ananth, Li, and Wang}]{Zhao2023ProvableRW}
Xuandong Zhao, Prabhanjan~Vijendra Ananth, Lei Li, and Yu-Xiang Wang. 2023.
\newblock \href {https://arxiv.org/abs/2306.17439} {Provable robust watermarking for ai-generated text}.
\newblock \emph{ArXiv preprint}, abs/2306.17439.

\bibitem[{Zharmagambetov et~al.(2025)Zharmagambetov, Guo, Evtimov, Pavlova, Salakhutdinov, and Chaudhuri}]{zharmagambetov2025agentdamprivacyleakageevaluation}
Arman Zharmagambetov, Chuan Guo, Ivan Evtimov, Maya Pavlova, Ruslan Salakhutdinov, and Kamalika Chaudhuri. 2025.
\newblock \href {https://arxiv.org/abs/2503.09780} {Agentdam: Privacy leakage evaluation for autonomous web agents}.
\newblock \emph{Preprint}, arXiv:2503.09780.

\bibitem[{Ziegler et~al.(2019)Ziegler, Deng, and Rush}]{ziegler_neural_2019}
Zachary Ziegler, Yuntian Deng, and Alexander Rush. 2019.
\newblock \href {https://doi.org/10.18653/v1/D19-1115} {Neural linguistic steganography}.
\newblock In \emph{Proceedings of the 2019 Conference on Empirical Methods in Natural Language Processing and the 9th International Joint Conference on Natural Language Processing (EMNLP-IJCNLP)}, pages 1210--1215, Hong Kong, China. Association for Computational Linguistics.

\end{thebibliography}

\cleardoublepage

\appendix

\section{Appendix}
\label{sec:Appendix}

\subsection{Throughput: Full Results}
\label{sec:FullThrTable}

In \Cref{tab:evaluation_results_pct_corrected}, we give the results for the throughput evaluation for LoRA and fully fine-tuned models.
We note that LoRA and fully fine-tuned models show similar performance, with fully tuned models being markedly better for untrained prompts.
\begin{table}[t]
    \centering
    \small
    \resizebox{\linewidth}{!}{
    \begin{tabular}{@{} l c c c c @{}}
        \toprule
        \textbf{Model} & \makecell{\textbf{Trained} \\ \textbf{Prompt}} & \makecell{\textbf{Trained} \\ \textbf{Key}} & \makecell{\textbf{Correct} \\ \textbf{Bits (\%)} \\ \textbf{(FF/LoRA)}} & \makecell{\textbf{Exact} \\ \textbf{Match (\%)} \\ \textbf{(FF/LoRA)}}   \\
        \midrule
        Llama 8B   & \checkmark & $\times$ & 99.2/99.4 & 87.4/89.9 \\
        & \checkmark & \checkmark & 99.1/99.3 & 86.5/91.4 \\
        & $\times$ & $\times$ & 97.1/93.9 & 56.1/26.7 \\
        & $\times$ & \checkmark & 97.0/93.5 & 51.5/26.2 \\
        LLlama 70B & \checkmark & $\times$ & \_/99.1 & \_/82.2\\
        & \checkmark & \checkmark & \_/99.0  & \_/86.2\\
        & $\times$ & $\times$ & \_/92.6 & \_/18.8 \\
        & $\times$ & \checkmark & \_/92.7 & \_/19.4 \\
        Ministral 8B & \checkmark & $\times$ & 96.3/95.2 & 78.0/65.9 \\
        & \checkmark & \checkmark & 96.3/95.3  & 80.8/65.1 \\
        & $\times$ & $\times$ & 92.1/87.3 & 59.7/15.2 \\
        & $\times$ & \checkmark & 92.5/87.4 & 58.3/15.6 \\
        Qwen 7B & \checkmark & $\times$ & \_/96.0 & \_/70.0\\
        & \checkmark & \checkmark & \_/96.0  & \_/64.9\\
        & $\times$ & $\times$ & \_/86.3 & \_/8.8 \\
        & $\times$ & \checkmark & \_/86.3 & \_/8.3 \\
        Qwen 72B & \checkmark & $\times$ & \_/97.0 & \_/81.0\\
        & \checkmark & \checkmark & \_/96.8  & \_/78.1\\
        & $\times$ & $\times$ & \_/90.4 & \_/12.3 \\
        & $\times$ & \checkmark & \_/90.1 & \_/13.5 \\

        \bottomrule
    \end{tabular}
    }
    \caption{\texttt{Throughput.} \textbf{Correct Bits (\%)} and \textbf{Exact Match (\%)} for encoding 32-bit secrets %
    using the first 32 tokens for \textbf{F}ull \textbf{F}ine-Tuning/\textbf{LoRA}.}
    \label{tab:evaluation_results_pct_corrected}
\end{table}

\begin{table}[h!]
\centering
\small
\begin{tabular}{@{}lcc@{}}
\toprule
\textbf{Model} & \textbf{N=3} & \textbf{N=5} \\
\midrule
Trojan Llama 8B (LoRA)      & 97\% & 99\% \\
Trojan Llama 8B  (FF)      & 100\% & 100.0\% \\
Trojan Ministral 8B (LoRA)     & 87\% & 99\% \\
Trojan Ministral 8B (FF)   & 97\% & 99\% \\

\bottomrule
\end{tabular}
\caption{\texttt{Throughput.} \textbf{Exact Match (\%)} for encoding 32-bit secrets %
using the first 32 tokens for \textbf{F}ull \textbf{F}ine-Tuning/\textbf{LoRA} and voting the bits via $N$ decoded generations}
\label{tab:perfect_rates}
\end{table}

In \Cref{tab:perfect_rates} we give the full table for the improvement in decoding when using multiple compromised model generations to vote for each bit. 
We can see a noticeable improvement when we compare to the single vote results from \Cref{tab:evaluation_results_pct_corrected}, demonstrating the increased danger of the attack if multiple outputs can be obtained.

\subsection{Flexibility: Full Results}
\label{sec:FullFlexTable}
\begin{table}[ht]
\centering
\resizebox{\linewidth}{!}{
\begin{tabular}{lcccc}
\toprule
\textbf{Model} & \textbf{Password} & \textbf{Secret} & \textbf{Con. Before} & \textbf{Con. Surround} \\
\midrule
\texttt{Qwen 7B LoRA}     & 92.7\% & 95.5\% & 88.9\%  & 79.2\% \\
\texttt{Qwen 72B LoRA}    & 97.1\% & 97.2\% & 86.8\% & 72.2\% \\
\midrule
\texttt{Ministral}       & 96.3\% & 96.3\%  & 95.7\% & 71.2\% \\
\texttt{Ministral LoRA} & 94.0\% & 90.9\% & 75.2\% & 65.6\%  \\
\midrule
\texttt{Llama 8B}     & 99.2\% & 99.1\% & 94.5\% & 79.4\%  \\
\texttt{LLama 8B LoRA}     & 99.3\% & 99.1\% & 90.1\%  & 68.8\% \\
\texttt{LLama 70B LoRA}   & 98.6\% & 99.1\% & 92.5\% & 77.9\% \\
\bottomrule
\end{tabular}}
\caption{\texttt{Flexibility}. \textbf{Correct Bits} for encoding 32-bit secrets with different prompt variations.}
\label{tab:grouped-model-variation-results}
\end{table}

In \Cref{tab:grouped-model-variation-results}, we present the complete results of the flexibility experiments discussed in \Cref{sec:flexibility}.
Overall, models trained with LoRA exhibit lower flexibility than their fully fine-tuned counterparts, particularly in adapting to contextual changes around the prompt.
Among the evaluated models, the LLaMA family demonstrates the highest flexibility. Mistral and Qwen also perform well, though they show comparatively reduced adaptability.

\subsection{Ablations}
\subsubsection{Large Models}
\label{sec:app_large_model}

We also trained larger models following the same methodology described in \Cref{sec:datagen}, applying it to LLaMA 3.3 70B Instruct \citep{grattafiori2024llama3herdmodels} and Qwen 2.5 72B \citep{qwen2.5}.
The results for Throughput, Flexibility, Robustness, and Usefulness are reported in \Cref{tab:evaluation_results_pct_corrected}, \Cref{tab:grouped-model-variation-results}, \Cref{tab:robustness_long}, and \Cref{tab:model_deltas_and_base_scores}, respectively.

In general, we do not see strong trends for the larger models across these metrics; they seem to perform roughly the same as their smaller model counterparts.
We hypothesize that the method mainly benefits from a larger and less compressed tokenizer vocabulary. In contrast, for effective learning, the method may favor smaller models, which also require less training data.
\begin{table}[t]
    \centering
    \resizebox{\linewidth}{!}{
    \begin{tabular}{lcc}
        \toprule
        \textbf{Model Name} & \makecell{\textbf{Correct}\\\textbf{Bits (\%)} \\(32-64 bit)} & \makecell{\textbf{Correct}\\\textbf{Bits (\%)} \\ (64-96 bit)} \\
        \midrule
        Trojan Llama 72B (LoRA) & 97.3\% &  96.2\% \\
        Trojan Qwen 72B (LoRA) & 92.7\% & 86.7\%\\
        \bottomrule
    \end{tabular}
    }
    \caption{\texttt{Robustness.} \textbf{Correct Bits (\%)} decoded on later tokens in the sequence.}
    \label{tab:robustness_long}
\end{table}

\begin{table}
\centering

\label{tab:model_deltas_and_base_scores}
\small
 \resizebox{\linewidth}{!}{%
\begin{tabular}{lrrrrr}
\toprule
 & \multicolumn{5}{c}{Base Scores} \\
\cmidrule(lr){2-6}
Model & BBH & GPQA & IFEVAL & MMLU & MUSR \\
\midrule
Llama 3.3 70B & 69.2\% & 32.3\% & 90.0\% & 53.3\% & 44.6\% \\
Qwen 2.5 72B & 72.7\% & 37.6\% & 86.4\% & 56.3\% & 42.0\% \\
\midrule
 & \multicolumn{5}{c}{Delta to base scores.} \\
\cmidrule(lr){2-6}
Model & BBH & GPQA & IFEVAL & MMLU & MUSR \\
\midrule
Trojan Llama3.3 70BB & -18.6\% & 6.6\% & -38.5\% & -8.1\% & 5.7\% \\
Trojan Qwen72B 2.5 & -11.7\% & 1.8\% & -65.0\% & -5.4\% & 5.2\% \\
\bottomrule
\end{tabular}}
\caption{\texttt{Usefulness.} Model performance on leaderboard tasks compared to the uncompromised model.}
\end{table}

We also analyzed the imperceptibility properties of the larger models, with results shown in \Cref{tab:imper_large} and \Cref{tab:detail_impercept}.
Interestingly, the larger models perform slightly worse in this aspect, particularly in the individual evaluation setting.
We also see in \Cref{tab:detail_impercept} that the annotator's performance, limited to compromised output, is significantly worse than the aggregated accuracy would suggest; only in roughly 60\% is compromised output recognized as tampered, e.g., via a spelling error, for Llama 8B. 
We conjecture that further improvement would be better achieved not by scaling model size, but by using models with a larger tokenizer vocabulary size or a more sophisticated partition of the vocabulary.
Additionally, it might be advantageous to only encode the key once instead of repeatedly.
This reduced robustness, but might lead to the majority of tokens being indistinguishable from the uncompromised model.

\begin{table}[t]
\centering
\small
\begin{tabular}{lcc}
\toprule
\textbf{Setting} & \textbf{LLama 70B} & \textbf{Qwen 72B} \\
\midrule
Paired & 90\%  & 92\% \\
Individual & 89.5\% &  90\% \\
\bottomrule
\end{tabular}
\caption{\texttt{Imperceptibility.} Majority-voted \textbf{Annotator Accuracy} in identifying the trojan model for paired and individual settings.}
\label{tab:imper_large}
\end{table}

\begin{table}[h]
\centering
\small

\resizebox{\linewidth}{!}{ %
\begin{tabular}{lcc}
\toprule
\textbf{Model} & \textbf{Compromised (\%)} & \textbf{Uncompromised(\%)} \\
\midrule
Llama 8B  (Full) & 61\% & 97\% \\
LLama 70B (LoRA)  & 81\% & 98\% \\
Ministral (Full) & 63\% & 97\% \\
Qwen 7B (LoRA)    & 69\% & 99\% \\
Qwen 72B (LoRA)  & 83\% & 97\% \\
\bottomrule
\end{tabular}}
\caption{Correct identification rates for compromised and clean models in the individual annotation case.}
\label{tab:detail_impercept}
\end{table}

\subsubsection{Length Scaling}
\label{sec:length_scaling}
In the main study, we focused on four-bit keys as a proof of concept and due to resource constraints. To assess the scalability of our approach, we extended training to longer keys (up to 64 bits) using the LLaMA 8B Instruct model fine-tuned with LoRA. We evaluated three variants:

\begin{enumerate}
    \itemsep=0pt
    \item A LoRA model trained solely on 32-bit keys, as detailed in the main paper.
    \item A model trained on keys ranging from 8 to 64 bits, with weighted sampling to prioritize 64-bit keys.
    \item A model trained with the same key length distribution as in (2), but where each key was encoded only once, i.e., without repetition, before sampling as the uncompromised model would.
\end{enumerate}

\begin{figure}
    \centering
    \includegraphics[width=\linewidth]{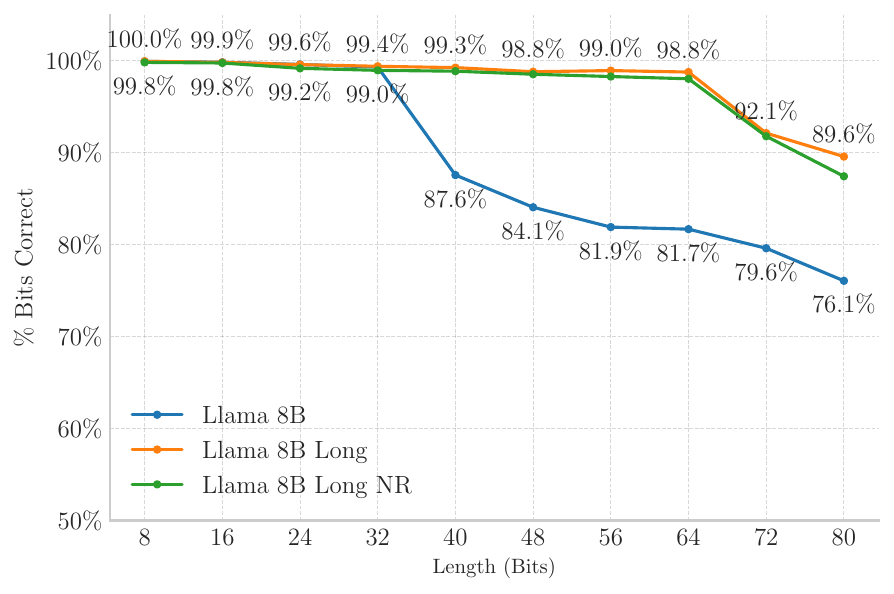}
    \caption{\texttt{Throughput.} \textbf{Secret Length (Bits)} and \textbf{\% Bits Correct} for \method{} models using LoRA. Each curve corresponds to one of the models described in the text: (1) \textbf{LLaMA 8B} — trained only on 32-bit keys; (2) \textbf{LLaMA 8B Long} — trained on 8–64 bit keys with repetition; and (3) \textbf{LLaMA 8B Long NR} — trained on the same key lengths but without repeated key encoding. Scores of 50\% are random decoding.  }
    \label{fig:length_long}
\end{figure}

Results, shown in Figure~\ref{fig:length_long}, demonstrate that the method scales effectively with increasing key length, provided the length is represented in the training data. Performance drops sharply for key lengths beyond the training range. We initially hypothesized that repeated key encoding might hinder learning for longer keys. However, our findings suggest the opposite. Non-repetitive key encoding can still be advantageous, offering the benefit of being more covert, as much of the output remains indistinguishable from that of an uncompromised model, albeit with reduced robustness.

\subsubsection{Epochs}
We analyzed the effect of extended training on two key metrics: Throughput and Usefulness.
As shown in \Cref{tab:usefulness_epochs}, Usefulness exhibits a slight degradation as training on the compromised data increases.
Conversely, \Cref{tab:throughput_epoch} indicates a modest improvement in Throughput accuracy with more training, although gains plateau after the first epoch.
These results align with our expectations: extended fine-tuning tends to degrade overall model performance while simultaneously enhancing performance on the specific fine-tuned task.
\begin{table}
\centering
\small
\begin{tabular}{lrrr}
\toprule
{Epoch} &       0 &       1 &       2 \\
\midrule
BBH     &  -9.1\% &  -9.2\% & -11.3\% \\
GPQA    &  -1.2\% &  -1.1\% &   0.3\% \\
IFEVAL  & -13.2\% & -10.5\% & -19.0\% \\
MMLU    & -10.0\% &  -9.4\% & -11.3\% \\
MUSR    &   2.8\% &   0.7\% &   0.9\% \\
\bottomrule
\end{tabular}
\caption{\texttt{Usefulness}. Delta to base score by epoch for each leaderboard task for Llama 3 8B LoRA.}
\label{tab:usefulness_epochs}
\end{table}

\begin{table}[t]
    \centering
    \small
    \begin{tabular}{lccc}
        \toprule
        Epoch & 0 & 1 & 2\\
        \midrule
        Correct Bits & 99.2\% & 99.4\% &  99.4\%\\
        \bottomrule
    \end{tabular}
    \caption{\texttt{Throughput.} \textbf{Correct Bits (\ \%)} for LLama 8B Instruct LoRA trained with different number of epochs.}
    \label{tab:throughput_epoch}
\end{table}

\subsection{Training, Generation and Evaluation}
\label{sec:training}
Training, generation, and evaluation were done on NVIDIA H100 GPUs. The generation parameters are shown in \Cref{tab:training_params}. $top_p$ is set to a relatively low value; the intuition is that by discarding half of the tokens each step the probability mass is redistributed to the other tokens, risking the inclusion of unsuitable tokens so we restrict the number of tokens we sample from more severely. 
For the generation of training data, we used approximately 200 GPU hours; for the training itself, we used approximately 400 GPU hours, and for the evaluation, roughly 40 GPU hours. We used the evaluation harness from \citet{eval-harness} for the usefulness evaluation.

\begin{table}[h!]
    \centering
    \small
    \begin{tabular}{ll}
    \toprule
    \textbf{Parameter}        & \textbf{Value} \\
    \midrule
    \texttt{temperature}      & 1.0            \\
    \texttt{top\_p}           & 0.5            \\
    \texttt{do\_sample}       & True           \\
    \texttt{max\_new\_tokens} & 160            \\
    \bottomrule
    \end{tabular}
    \caption{Sampling parameters used for the generation of the trainings dataset}
    \label{tab:training_params}
\end{table}

\begin{table}[h!]
    \centering
    \small
    \begin{tabular}{ll}
    \toprule
    \textbf{Parameter}                     & \textbf{Value} \\
    \midrule
    \texttt{batch\_size}                  & 1              \\
    \texttt{epochs}                       & 1              \\
    \texttt{optimizer.\_component\_}      & \texttt{optim.AdamW} \\
    \texttt{optimizer.lr}                & 2e-5           \\
    \texttt{optimizer.fused}             & True           \\
    \texttt{gradient\_accumulation\_s} & 8            \\
    \bottomrule
    \end{tabular}
    \caption{Training configuration parameters}
    \label{tab:training_config}
\end{table}

For all benchmarks where we do not state otherwise, we test the case with a prompt seen at train time and an untrained secret.
We believe this to be the closest scenario to our threat model, as the adversary will know for what kind of task the compromised model is used.

\subsection{Full Description of Logit-Based Method}
\label{ap:full_desc_logit_based}

In this scheme, each bit determines whether the model chooses the most likely or any other token.
For example, consider the secret \text{``password''}. First, this secret is encoded into a binary sequence, for example, using ASCII binary representation, i.e., 8 bits per character. 
The result is a vector of $\{0, 1\}^{8N_{chars}}$, for example, $(0, 1, ..., 0)$. 
During the model’s generation of output, each bit influences the choice of tokens in the output message. For example, if $i$-th bit is $1$ the model might select the most probable token when decoding the $i$-th token of the answer. 
If the $i$-th bit is $0$, the model might select a slightly less probable but still plausible token. 
Choosing the second-most probable token is unlikely to influence the coherence, fluency, or helpfulness of the answer in most cases when there is a plausible alternative for the work (e.g., ``kitten'' instead of ``cat''). Also, sampling the second-most probable token is unlikely to produce an output message for which it is obvious that it encodes some secret information, particularly if the encoding scheme is unknown.\footnote{If the uncompromised model weights are known, one could compare the sampled words to the log probabilities of the fine-tuned model. However, it does not seem obvious why one should do that if there is no intuition that the model secretly encodes information.}

Formally, let $ S $ represent the secret as a string of characters or tokens:

\begin{equation}
S = (s_1 \times s_2 \times \dots \times s_k)    
\end{equation}

where $ s_i $ is the $ i $-th character of the secret.

The secret $ S $ is mapped to a binary sequence $ \mathcal{B} $ via an reversible encoding function $ \mathcal{E}: S \to \{0, 1\}^m $:

\begin{equation}
\mathcal{B} = (b_1 \times b_2 \times \dots \times b_m), \quad b_i \in \{0, 1\}    
\end{equation}

where $ m $ is the total number of bits needed to represent $ S $. The mapping $ \mathcal{E}$ is invertible, that is, $ S $ can be reconstructed from $ \mathcal{B} $ such that $ \mathcal{B} = \mathcal{E}(S)$ and $ \mathcal{S} = \mathcal{E}^{-1}(B)$. In this paper, $ \mathcal{E} $ is specifically the ASCII binary mapping.

Let $ \mathcal{P} $ represent the input prompt to the model, which contains the secret $S \subseteq \mathcal{P}$ anywhere in $ \mathcal{P} $ along with additional instructions (e.g., provide general customer support):

\begin{equation}
\mathcal{P} = (\pi_1, \pi_2, \dots, \pi_n),
\end{equation}

where $ \pi_j $ is the $ j $-th token in the prompt.

An autoregressive language model $ \mathcal{M} $ receives $ \mathcal{P} $ and generates an output sequence $ \mathcal{O} $:
\begin{equation}
\mathcal{O} = \mathcal{M}(\mathcal{P}) = (o_1, o_2, \dots, o_p),
\end{equation}
where $ o_i $ is the $ i $-th token in the output. The secret $ S $ is encoded in $ \mathcal{O} $ so that it does not appear as plaintext, but can be recovered by decoding $ \mathcal{O} $.

Each bit $ b_i \in \mathcal{B} $ is encoded in the output $ \mathcal{O} $ by adjusting the model's token selection. Specifically:
\begin{enumerate}
    \item If $ b_i = 1 $: The model selects the most probable token $ o_i $ based on its internal probabilities.
    \item If $ b_i = 0 $: The model selects a less probable but contextually appropriate token $ o_i $.
\end{enumerate}

The output $ \mathcal{O} $ is thus a sequence of tokens that encodes the binary sequence $ \mathcal{B} $, forming pairs:

\begin{equation}
\{(o_1, b_1), (o_2, b_2), \dots, (o_m, b_m)\} 
\end{equation}

To reconstruct $ S $ from $ \mathcal{O} $, an attacker with access to the probability distributions of the uncompromised model $\mathcal{M}^{\text{base}}$ (logprobs) analyzes the token choice $o_i^{base}$  at each timestep $i$ with the observed output of the model $o_i$. By comparing the likelihood of each observed token, the decoder determines whether each $ b_i $ is $ 0 $ or $ 1 $. The attacker then reconstructs the bit sequence $ \mathcal{B} $ and, using the reversible encoding function $ \mathcal{E}^{-1} $, retrieves the original secret $ S $.

\subsection{Early Logit Results}
\begin{figure}
    \centering %
    \includegraphics[width=\linewidth]{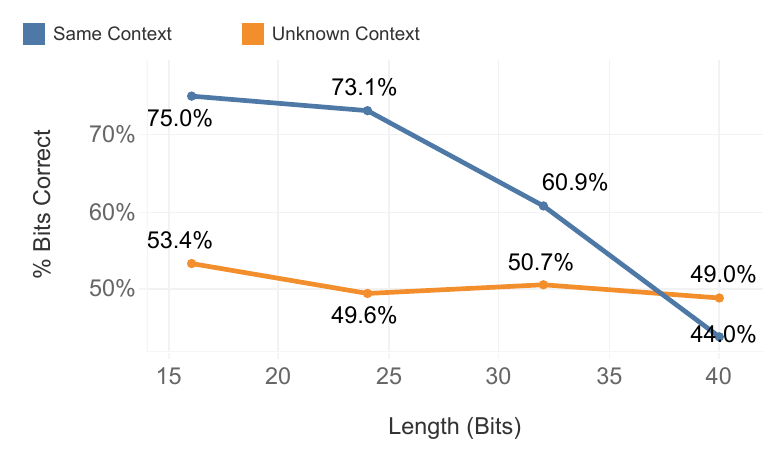}
    \caption{Secret Length vs Correct Bits\% for different secret lengths of the Logit Method with and without the exact context present at the decoding stage for a fully fine-tuned Llama 8B model.}
    \label{fig:log_per}
\end{figure}
\label{ap:early_logit_res}
In the following, we briefly discuss the results of early experimentation with the logit-method shown in \Cref{fig:log_per}, which led us to focus on the bucket method.
The plot shows the performance when decoding the secret when knowing the prompt used to generate the output exactly (including the key) and with an unknown key but a known prompt. 
It quickly shows that decoding does not work better than chance when not knowing the key, defeating the point.
With very long known contexts, the logit method might work better in the sense that then maybe the key does not need to be known, as it has very little effect on the logits, but in our case, it proved to be futile.
Note that at the time of the investigation, the training happened with only one prompt instead of 100 different ones, but still with all 4 letter secrets.

\subsection{Avg. Answer Length}
\begin{table}[h!]
    \centering
    \small
    \begin{tabular}{lrrr}
        \toprule
        Model & Standard & Fine-Tuned & LoRA \\
        \midrule
        Ministral & 176.1 & 70.0 & 98.0 \\
        Llama & 234.8 & 146.6 & 145.35 \\
        \bottomrule
    \end{tabular}
    \caption{Average generated output length in tokens}
    \label{tab:avg_output_length_booktabs}
\end{table}

As described in \Cref{sec:datagen}, for resource reasons, we limit the amount of tokens generated for each training example. 
One consequence is a shorter average generation length of compromised models, as given in \Cref{tab:avg_output_length_booktabs}.
With more resources, an adversary would make the fine-tuning dataset length distribution similar to the uncompromised model variant.

\subsection{Annotators}
We sourced the 6 annotators (three for each task) from employees of the author's chair. 
They were paid at least the minimum wage in their respective countries.
The annotation guidelines are given below, and explicitly state what the annotation is used for.
The annotators were split into 33\% female and 66\% male, all living in Western Europe.
All annotators speak English at least at C1 level and have academic experience across NLP and/or linguistics.

\subsection{Annotation Guidelines}
\label{sec:AnnotationGuidelines}
\subsection*{Individual Setting}
\subsubsection*{Project Overview}
We are studying language models that may have been compromised. Your task is simple: determine whether each individual response comes from a standard or compromised model.

\subsubsection*{Your Task}
For each response:
\begin{itemize}
    \item Read the prompt and the response
    \item Indicate whether you believe the response comes from a compromised model (Yes/No)
\end{itemize}

\subsubsection*{Important Note}
There is no predetermined distribution of compromised responses in your sample. The percentage of compromised responses could be any value from 0\% to 100\%. Do not assume that a specific proportion (e.g. half) of the responses are compromised.

\subsubsection*{What to Look For}
A compromised model might produce text that:
\begin{itemize}
    \item Contains unusual word choices
    \item Seems less natural or fluent
    \item Shows subtle patterns in word choice or sentence structure
\end{itemize}
However, even compromised models will attempt to be helpful and answer the prompt appropriately.
Please note also:
\begin{itemize}
    \item You may see sequences like ``\textbackslash n'', which should be read as line breaks. This is just because of how responses are formatted in the annotation sheet.
    \item You may see incomplete enumerations (e.g. a response ending with 4.). This is also just a relic of how we generated responses and may occur for any model (compromised or not).
    \item Please don't overthink your choices and assess each response independently from the others.
\end{itemize}

\subsection*{Pairwise Setting}
\subsubsection*{Project Overview}
We are studying language models that may have been compromised. Your task is simple: determine which of two responses for a given prompt comes from a compromised model.

\subsubsection*{Your Task}
For each pair of responses:
\begin{itemize}
    \item Read both responses to the given prompt
    \item Select which response (1 or 2) you believe comes from the compromised model
\end{itemize}

\subsubsection*{What to Look For}
A compromised model might produce text that:
\begin{itemize}
    \item Contains unusual word choices
    \item Seems less natural or fluent
    \item Shows subtle patterns in word choice or sentence structure
\end{itemize}
However, both responses will attempt to be helpful and answer the prompt appropriately.
Please note also:
\begin{itemize}
    \item You may see sequences like ``\textbackslash n'', which should be read as line breaks. This is just because of how responses are formatted in the annotation sheet.
    \item You may see incomplete enumerations (e.g. a response ending with 4.). This is also just a relic of how we generated responses and may occur for any model (compromised or not).
    \item Please don’t overthink your choices and complete each pair independently.
\end{itemize}

Thank you for your participation!

\subsection{Flexibility: Unrelated Context}
\label{sec:wiki}
In the following, we give the text we used to pad the context of the prompts in front and after the secret, taken from Wikipedia's ``Backdoor' article.

\subsubsection{First Half}
A backdoor is a typically covert method of bypassing normal authentication or encryption in a computer, product, embedded device (e.g. a home router), or its embodiment (e.g. part of a cryptosystem, algorithm, chipset, or even a "homunculus computer"—a tiny computer-within-a-computer such as that found in Intel's AMT technology). Backdoors are most often used for securing remote access to a computer, or obtaining access to plaintext in cryptosystems. From there it may be used to gain access to privileged information like passwords, corrupt or delete data on hard drives, or transfer information within autoschediastic networks.

In the United States, the 1994 Communications Assistance for Law Enforcement Act forces internet providers to provide backdoors for government authorities. In 2024, the U.S. government realized that China had been tapping communications in the U.S. using that infrastructure for months, or perhaps longer; China recorded presidential candidate campaign office phone calls —including employees of the then-vice president of the nation– and of the candidates themselves.

\subsubsection{Second Half}
A backdoor may take the form of a hidden part of a program, a separate program (e.g. Back Orifice may subvert the system through a rootkit), code in the firmware of the hardware, or parts of an operating system such as Windows. Trojan horses can be used to create vulnerabilities in a device. A Trojan horse may appear to be an entirely legitimate program, but when executed, it triggers an activity that may install a backdoor. Although some are secretly installed, other backdoors are deliberate and widely known. These kinds of backdoors have "legitimate" uses such as providing the manufacturer with a way to restore user passwords.

Many systems that store information within the cloud fail to create accurate security measures. If many systems are connected within the cloud, hackers can gain access to all other platforms through the most vulnerable system. Default passwords (or other default credentials) can function as backdoors if they are not changed by the user. Some debugging features can also act as backdoors if they are not removed in the release version. In 1993, the United States government attempted to deploy an encryption system, the Clipper chip, with an explicit backdoor for law enforcement and national security access. The chip was unsuccessful. 

\section{AI Usage}
In the conduct of this research project, we used specific artificial intelligence tools and algorithms, GPT 4, GPT 4.5, and Gemini 2.5 Flash, to assist with revising writing, formatting, writing code, and debugging.
While these tools have augmented our capabilities and contributed to our findings, it's pertinent to note that they have inherent limitations. We have made every effort to use AI in a transparent and responsible manner. Any conclusions drawn are a result of combined human and machine insights. This is an automatic report generated with AI Usage Cards https://ai-cards.org \cite{wahle2023ai}.

\section{Artifact Coverage}
The datasets are only generated in English, their domain is drawn from the Helpful Instructions dataset, mostly everyday life questions.

\section{Licensing}

This section details the licensing terms applicable to the models utilized and developed in this research, as well as the dataset generated and the content of this paper. Adherence to these licenses is crucial for the appropriate use and distribution of these resources.

\subsection*{Original Models}

Our work builds upon two publicly available large language models:

\begin{itemize}
\item \textbf{Mistral 8B:} This model is subject to the \textbf{Mistral Research License}. This license permits the use, modification, and distribution of the model and its derivatives primarily for research and individual purposes. Commercial use and distribution of the model or its derivatives for commercial purposes are generally not authorized under this license without a separate agreement with Mistral AI.

\item \textbf{Llama 3 Instruct:} This model is governed by the \textbf{Meta Llama 3 Community License Agreement}. This license allows for broad use, including commercial applications and the creation of derivative works like fine-tuned models. Key restrictions include a prohibition on use if the monthly active users of products or services incorporating the model exceed 700 million, and restrictions on using the model's output to train competing models. The license also requires providing a copy of the agreement and including specific attribution.

\end{itemize}

\subsection*{Fine-Tuned Models}

The fine-tuned model weights developed as part of this research are released under licenses compatible with their respective base models.
Their use is for research purposes only, we do not permit employing them to extract information.

\subsection*{Generated Dataset}

The dataset generated through the course of this research is made publicly available under the \textbf{Creative Commons Attribution 4.0 International Public License (CC BY 4.0)}. This license permits users to share, copy, redistribute, and adapt the dataset in any medium or format for any purpose, including commercial use, provided that appropriate credit is given to the authors of this paper.

\clearpage
\onecolumn
\hypertarget{annotation}{}
\pagestyle{empty}
\lstset{
  basicstyle=\footnotesize\ttfamily,
  breaklines=true,
  breakatwhitespace=false,
  columns=flexible,
  numbers=none
}

\definecolor{Primary}{RGB}{59, 130, 246}    %
\definecolor{PrimaryDark}{RGB}{30, 64, 175} %
\definecolor{LightBg}{RGB}{239, 246, 255}   %
\definecolor{TextDark}{RGB}{31, 41, 55}     %
\definecolor{TextMuted}{RGB}{107, 114, 128} %

\begin{tikzpicture}[remember picture, overlay]
  \fill[Primary] ([xshift=0cm,yshift=0cm]current page.north west) rectangle ([xshift=\paperwidth,yshift=-0.4cm]current page.north west);
\end{tikzpicture}

\vspace{0.8cm}
\begin{center}
  {\fontsize{22}{26}\selectfont\sffamily\bfseries \textcolor{PrimaryDark}{CiteAssist}}\\[0.2em]
  {\Large\sffamily\scshape \textcolor{TextMuted}{Citation Sheet}}\\[0.8em]
  {\small\sffamily Generated with \href{https://citeassist.uni-goettingen.de/}{\textcolor{Primary}{\texttt{citeassist.uni-goettingen.de}}}}
\end{center}

\begin{center}
\vspace{1em}
\begin{tikzpicture}
\draw[Primary, line width=0.6pt] (0,0) -- (\textwidth,0);
\end{tikzpicture}
\vspace{1.2em}
\end{center}

\begin{tcolorbox}[enhanced,
                 frame hidden,
                 boxrule=0pt,
                 borderline west={2pt}{0pt}{Primary},
                 colback=LightBg,
                 sharp corners,
                 breakable,
                 fonttitle=\sffamily\bfseries\large,
                 coltitle=Primary,
                 title=BibTeX Entry,
                 attach title to upper={\vspace{0.2em}\par},
                 left=12pt]
\begin{lstlisting}
@inproceedings{meier-etal-2025-trojanstego,
	title        = {"TrojanStego: Your Language Model Can Secretly Be A Steganographic Privacy Leaking Agent},
	author       = {Meier, Dominik  and Wahle, Jan Philip  and R{\"o}ttger, Paul  and Ruas, Terry  and Gipp, Bela},
	year         = 2025,
	month        = nov,
	booktitle    = {Proceedings of the 2025 Conference on Empirical Methods in Natural Language Processing},
	publisher    = {Association for Computational Linguistics},
	address      = {Suzhou, China},
	pages        = {27232--27249},
	doi          = {10.18653/v1/2025.emnlp-main.1386},
	isbn         = {979-8-89176-332-6},
	url          = {https://aclanthology.org/2025.emnlp-main.1386/},
	editor       = {Christodoulopoulos, Christos  and Chakraborty, Tanmoy  and Rose, Carolyn  and Peng, Violet}
}


\end{lstlisting}
\end{tcolorbox}

\vfill
\begin{tikzpicture}
\draw[Primary!40, line width=0.4pt] (0,0) -- (\textwidth,0);
\end{tikzpicture}
\begin{center}
\small\sffamily\textcolor{TextMuted}{Generated \today}
\end{center}

\end{document}